\documentclass[sigconf]{acmart}
\let\widebar\relax
\usepackage{mathabx}
\usepackage{amsfonts}
\usepackage{graphicx}
\usepackage{xcolor}
\usepackage{hyperref}
\usepackage{setspace}
\usepackage{booktabs}
\usepackage{subfigure}
\usepackage{color}
\usepackage[normalem]{ulem}
\usepackage{enumitem}
\usepackage{multirow}
\usepackage{ulem}
\usepackage[nomargin,inline,marginclue,draft]{fixme}
\usepackage{balance}
\usepackage{verbatim}
\usepackage{diagbox}
\usepackage{changepage}
\usepackage{bm}
\usepackage{amssymb}
\usepackage{xspace}
\usepackage{algorithm2e}
\usepackage{pifont}
\usepackage{soul}

\definecolor{myred}{rgb}{1, 0, 0}
\definecolor{myblue}{rgb}{0, 0, 1}
\definecolor{myblack}{rgb}{1, 1, 1}

\newcommand{\Ours}{Heterformer\xspace}
\newcommand{\tl}{textless\xspace}
\newcommand{\Tl}{Textless\xspace}

\newcommand{\bmh}{{\bm h}}
\newcommand{\bmW}{{\bm W}}
\newcommand{\bmz}{{\bm z}}
\newcommand{\bmH}{{\bm H}}

\newcommand{\bmQ}{{\bm Q}}
\newcommand{\bmK}{{\bm K}}
\newcommand{\bmV}{{\bm V}}
\definecolor{myblue}{rgb}{0.2, 0.2, 0.9}

\newcommand\cls{\texttt{[CLS]}\xspace}

\newlength\savedwidth

\copyrightyear{2023}
\acmYear{2023}
\setcopyright{acmlicensed}
\acmConference[KDD '23] {Proceedings of the 29th ACM SIGKDD Conference on Knowledge Discovery and Data Mining}{August 6--10, 2023}{Long Beach, CA, USA.}
\acmBooktitle{Proceedings of the 29th ACM SIGKDD Conference on Knowledge Discovery and Data Mining (KDD '23), August 6--10, 2023, Long Beach, CA, USA}
\acmPrice{15.00}
\acmISBN{979-8-4007-0103-0/23/08}
\acmDOI{10.1145/3580305.3599376}

\begin{document}

\title{Heterformer: Transformer-based Deep Node Representation Learning on Heterogeneous Text-Rich Networks}



\author{Bowen Jin}
\affiliation{
\institution{University of Illinois at Urbana-Champaign} 
\country{}
\institution{bowenj4@illinois.edu}
}

\author{Yu Zhang}
\affiliation{
\institution{University of Illinois at Urbana-Champaign} 
\country{}
\institution{yuz9@illinois.edu}
}

\author{Qi Zhu}
\affiliation{
\institution{University of Illinois at Urbana-Champaign} 
\country{}
\institution{qiz3@illinois.edu}
}

\author{Jiawei Han}
\affiliation{
\institution{University of Illinois at Urbana-Champaign} 
\country{}
\institution{hanj@illinois.edu}
}

\begin{abstract}
Representation learning on networks aims to derive a meaningful vector representation for each node, thereby facilitating downstream tasks such as link prediction, node classification, and node clustering.
In heterogeneous text-rich networks, this task is more challenging due to (1) \textit{presence or absence of text}: Some nodes are associated with rich textual information, while others are not; (2) \textit{diversity of types}: Nodes and edges of multiple types form a heterogeneous network structure.
As pretrained language models (PLMs) have demonstrated their effectiveness in obtaining widely generalizable text representations, a substantial amount of effort has been made to incorporate PLMs into representation learning on text-rich networks.
However, few of them can jointly consider heterogeneous structure (network) information as well as rich textual semantic information of each node effectively.
In this paper, we propose \Ours, a \textit{\textbf{Heter}ogeneous Network-Empowered Trans\textbf{former}} that performs contextualized text encoding and heterogeneous structure encoding in a unified model.
Specifically, we inject heterogeneous structure information into each Transformer layer when encoding node texts.
Meanwhile, \Ours is capable of characterizing node/edge type heterogeneity and encoding nodes with or without texts.
We conduct comprehensive experiments on three tasks (\textit{i.e.}, link prediction, node classification, and node clustering) on three large-scale datasets from different domains, where \Ours outperforms competitive baselines significantly and consistently.
The code can be found at 

\noindent\url{https://github.com/PeterGriffinJin/Heterformer}.

\end{abstract}



\begin{spacing}{0.96}
\maketitle

\renewcommand{\thefootnote}{\fnsymbol{footnote}}


\section{Introduction}\label{sec::intro}
\begin{figure}
\centering
\includegraphics[width=0.45\textwidth]{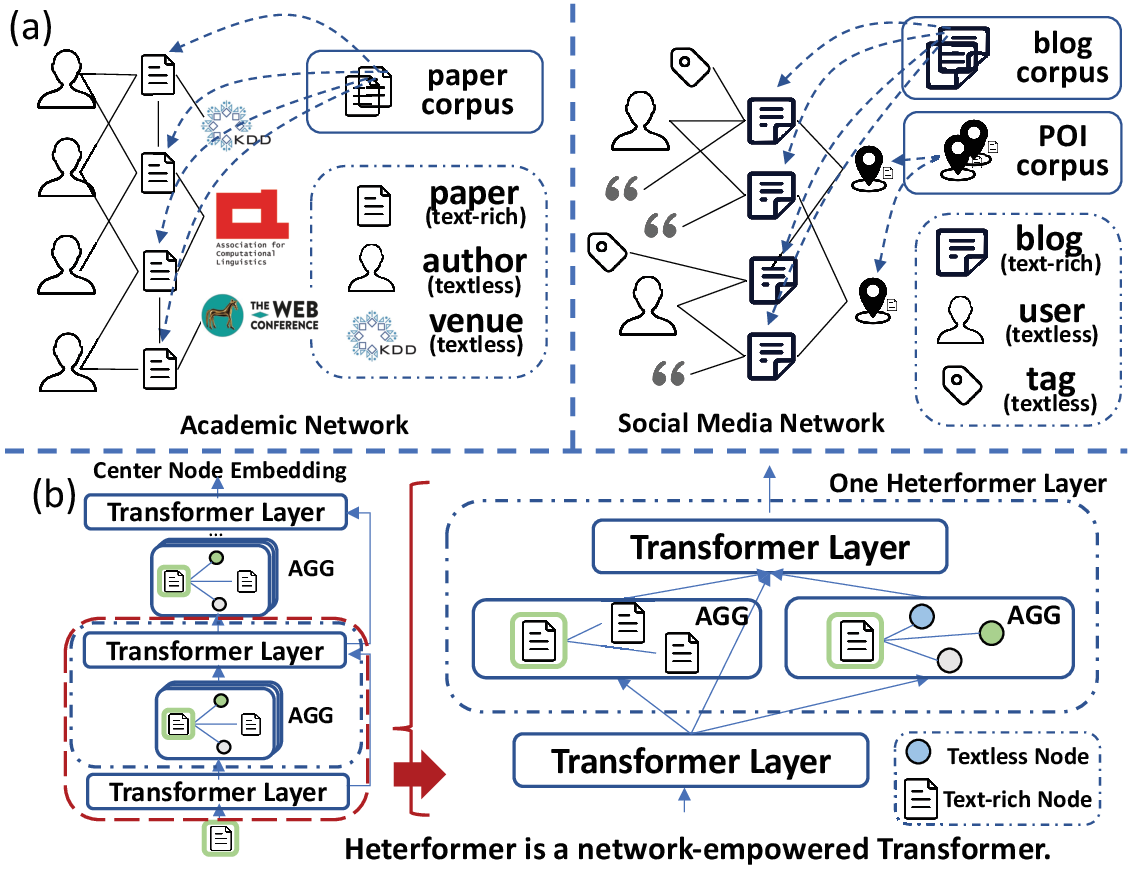}
\vspace{-0.3cm}
\caption{(a) Examples of heterogeneous text-rich networks: an academic network and a social media network. (b) An illustration of our heterogeneous network-empowered Transformer, \Ours. One \Ours layer is zoomed out. AGG denotes neighbor aggregation on the network.}
\label{fig:intro}
\end{figure}

Heterogeneous text-rich networks are ubiquitously utilized to model real-world data such as academic networks \cite{tang2008arnetminer}, product networks \cite{dong2020autoknow}, and social media \cite{el2022twhin}.
Such networks often have two characteristics: 
(1) \textit{Text-rich}: \textbf{some} types of nodes are associated with textual information. For instance, papers in academic networks \cite{tang2008arnetminer} have their titles and abstracts; tweets in social media networks \cite{el2022twhin} have their tweet contents.
(2) \textit{Heterogeneous}: nodes and edges in the network are multi-typed. For example, academic networks \cite{tang2008arnetminer} have paper, author, and venue nodes; product networks \cite{dong2020autoknow} have edges between users and products reflecting ``purchase'' and ``view'' relations. 
In such text-rich networks (Figure \ref{fig:intro}(a)), to obtain satisfying node representations which can be generalized to various tasks such as link prediction \cite{sun2011pathsim}, node classification \cite{wang2019heterogeneous}, and recommendation \cite{zhang2019heterogeneous}, the model needs to consider both \textbf{text semantics} and \textbf{heterogeneous} structure (network) information.

To capture the rich text semantic signals, Transformer \cite{vaswani2017attention} is a powerful architecture featured by its fully connected attention mechanism.
Taking Transformer as the backbone, pretrained language models (PLMs) \cite{devlin2018bert,liu2019roberta,clark2020electra} learned from web-scale corpora can obtain contextualized semantic representations of words and documents.
These representations are demonstrated to be of high quality in various text mining tasks \cite{sia2020tired,chang2020taming,liu2019fine}. 
To introduce the advanced capability of PLMs into node representation learning on text-rich networks, existing works mainly adopt a cascaded architecture \cite{jin2021bite,li2021adsgnn,zhu2021textgnn,zhang2019shne}, where the textual information of each node is first encoded via PLMs and then aggregated via graph encoders \cite{kipf2017semi, hamilton2017inductive, velivckovic2017graph}.
In such cases, the link connecting two nodes is not utilized when generating their text representations.
In fact, linked nodes can benefit each other regarding text semantics understanding. 
For example, the term ``Transformer'' in a paper cited by many machine learning papers should refer to a deep learning model rather than an electrical engineering component.
GraphFormers \cite{yang2021graphformers} further introduce a nested Transformer architecture for deep information integration between text encoding modules and network encoding modules. 
Patton \cite{jin2023patton} proposes two strategies to pretrain GraphFormers on text-rich networks.
Yet, they adopt a strict homogeneous network assumption (\textit{i.e.}, all the nodes are associated with semantically rich text and are of the same type), which is hard to be satisfied in practice.

As a matter of fact, real-world text-rich networks are usually heterogeneous with the heterogeneity coming from two sources: the presence or absence of text and the diversity of types. 
\vspace{3px}
\begin{itemize}[leftmargin=*,partopsep=0pt,topsep=0pt]
\setlength{\itemsep}{0pt}
\setlength{\parsep}{0pt}
\setlength{\parskip}{0pt}

\item \textbf{Presence or Absence of Text.}
Not every node in text-rich networks exhibits rich textual information.
Instead, some nodes are not guaranteed to contain textual information (\textit{e.g.}, many users are not associated with text information in social media networks).
Based on the presence or absence of text, nodes can be categorized into \textbf{text-rich} nodes  (associated with semantically rich text, \textit{e.g.}, tweets and papers) and \textbf{\tl} nodes (without semantically rich text, \textit{e.g.}, users and authors).
Text-rich nodes can intuitively contribute to representation learning with the rich texts, while \tl nodes can also be strong semantic indicators in the network. 
For example, given a paper node about ``Byzantine'' which is linked to an author node (\tl) with many paper neighbors related to ``distributed system'', 
we can infer that ``Byzantine'' here refers to a computer network term rather than an empire in history.
Since both text-rich nodes and \tl nodes should be considered, how to leverage both kinds of nodes in representation learning with PLMs is an open question that needs to be answered.

\item \textbf{Diversity of Types.} 
As previously stated, a large number of real-world networks contain nodes and edges of different types. 
For example, there are at least three types of nodes (``paper'', ``author'', and ``venue'' nodes) in academic networks \cite{tang2008arnetminer}; there are at least four types of edges (``click'', ``view'', ``cart'', and ``purchase'' edges) between users and products in e-commerce networks \cite{dong2020autoknow}.
Different types of nodes/edges have different traits and their features may fall in different feature spaces.
For instance, the feature of a user may contain gender, age, and nationality while the feature of an item may contain price and quality. 
How to handle such complex structural information while preserving the diverse type information simultaneously with PLMs is a crucial issue that needs to be solved.
\end{itemize}

\vspace{3px}
\noindent \textbf{Present Work.}
To this end, we propose a network-empowered Transformer (Figure \ref{fig:intro}(b)), \textit{i.e.}, \Ours, for node representation learning on heterogeneous text-rich networks, while capturing the two sources of heterogeneity mentioned above. 
Specifically:
(1) In \textit{the whole model}, we introduce virtual neighbor tokens inside each Transformer layer (initialized by the corresponding layer in a PLM, \textit{e.g.}, BERT \cite{devlin2018bert}) for text encoding, to fuse representations of each node’s text-rich neighbors, textless neighbors, and its own content via the fully connected attention mechanism. The virtual neighbor token hidden states are attention-based aggregations of neighbor node embeddings.
(2) To deal with the \textit{presence or absence of text}, two virtual neighbor tokens are utilized to capture the semantic signals from text-rich neighbors and \tl neighbors, respectively. Furthermore, we propose an embedding warm-up stage for \tl nodes to obtain better initial embeddings before the whole model training.
(3) To capture the \textit{diversity of types}, we use type-specific transformation matrices to project different types of nodes into the same latent space. When calculating virtual neighbor token hidden states, the aggregation module collects information from its neighbors by characterizing edge types in the attention mechanism.
The overall model is optimized via an unsupervised link prediction objective \cite{perozzi2014deepwalk, hamilton2017inductive}.

\vspace{3px}
The main contributions of our paper are summarized as follows:
\begin{itemize}[leftmargin=*,partopsep=0pt,topsep=0pt]
    \setlength{\itemsep}{0pt}
    \setlength{\parsep}{0pt}
    \setlength{\parskip}{0pt}
    \item We formalize the problem of node representation learning on heterogeneous text-rich networks, which involves joint encoding of heterogeneous network structures and textual semantics.
    \item We point out heterogeneity from two sources and propose a heterogeneous network-empowered Transformer architecture called \Ours, which deeply couples text encoding and heterogeneous structure (network) encoding.
    \item  We conduct comprehensive experiments on three public text-rich networks from different domains, where \Ours outperforms competitive baseline models (including GNN-cascaded Transformers and nested Transformers) significantly and consistently on various tasks, including link prediction, node classification, and node clustering.
\end{itemize}

\vspace{-0.2in}
\section{Preliminaries}\label{sec::profdef}
\subsection{Heterogeneous Text-rich Networks}

\noindent\textbf{Definition 2.1. Heterogeneous Networks \cite{sun2012mining}.} A heterogeneous network is defined as $\mathcal{G}=(\mathcal{V}, \mathcal{E}, \mathcal{A}, \mathcal{R})$, where $\mathcal{V}$, $\mathcal{E}$, $\mathcal{A}$, $\mathcal{R}$ represent the sets of nodes, edges, node types, and edge types, respectively. $|\mathcal{A}|+|\mathcal{R}|>2$. A heterogeneous network is also associated with a node type mapping function $\phi: \mathcal{V}\rightarrow\mathcal{A}$ and an edge type mapping function $\psi: \mathcal{E}\rightarrow\mathcal{R}$.

\vspace{3px}
\noindent\textbf{Definition 2.2. Text-Rich Nodes and \Tl Nodes.} In a heterogeneous network $\mathcal{G}=(\mathcal{V}, \mathcal{E}, \mathcal{A}, \mathcal{R})$, $v\in\mathcal{V}$ is \textbf{text-rich} if it is associated with semantically rich text information $doc\in\mathcal{D}$. $\mathcal{D}$ is the document set. Otherwise, it is \textbf{\tl}. We assume that nodes of the same type are either all text-rich or all \tl. 

\vspace{3px}
\noindent\textbf{Definition 2.3. Heterogeneous Text-Rich Networks \cite{zhang2019shne,shi2019discovering}.}
A heterogeneous network $\mathcal{G}=(\mathcal{V}, \mathcal{E}, \mathcal{A}, \mathcal{R}, \mathcal{D})$ is a heterogeneous text-rich network if $\mathcal{D}\neq \varnothing$, $\mathcal{A}=\mathcal{A}_\text{TR}\cup \mathcal{A}_\text{TL}$, $\mathcal{A}_\text{TR}\cap \mathcal{A}_\text{TL}=\varnothing$ and $\mathcal{A}_\text{TR} \neq \varnothing$, where $\mathcal{A}_\text{TR}$ and $\mathcal{A}_\text{TL}$ denote the sets of text-rich node types and \tl node types, respectively.

\begin{figure*}
\centering
\includegraphics[width=16cm]{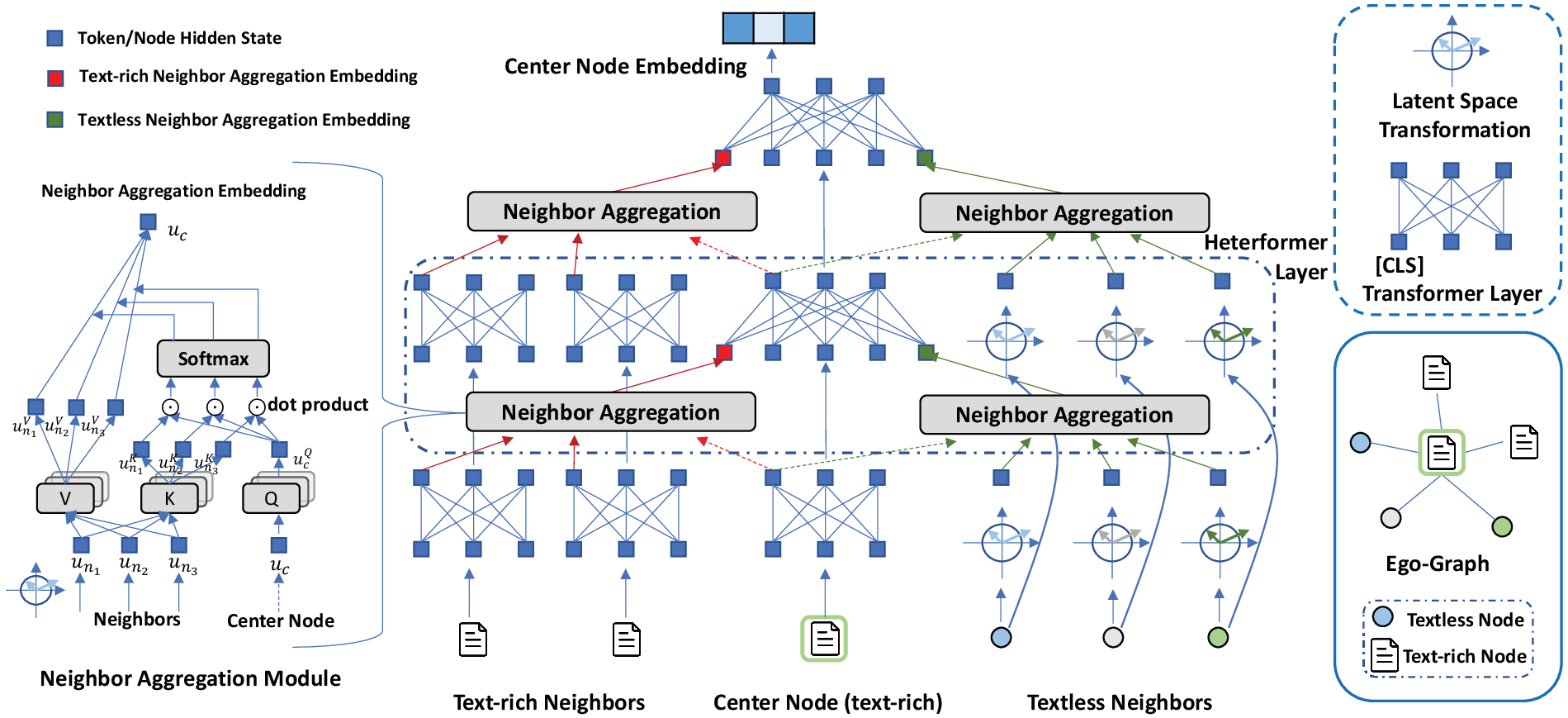}
\caption{The overall architecture of \Ours. There are two layers in the figure, while in experiments we have 11 layers. Different color denotes different types of nodes. The whole encoding procedure of \Ours can be found in Appendix \ref{apx::sec::alg}.}\label{fig::whole-structure}
\end{figure*}

\vspace{-0.2in}
\subsection{Transformer}
A large number of PLMs (\textit{e.g.}, BERT \cite{devlin2018bert}) utilize the multi-layer Transformer architecture \cite{vaswani2017attention} to encode texts. 
Each Transformer layer adopts a multi-head self-attention mechanism to gain a contextualized representation of each text token. Specifically, let $\bmH^{(l)}=[\bm{h}^{(l)}_1, \bm{h}^{(l)}_2, ..., \bm{h}^{(l)}_n]$ denote the output hidden states of the $l$-th Transformer layer, where $\bm{h}^{(l)}_i\in\mathcal{R}^d$ is the representation of the text token at position $i$.
Then, the multi-head self-attention (MHA) in the ($l$+1)-th Transformer layer is calculated as
\begin{gather}
    {\rm MHA}({\bmH}^{(l)}) = \mathop{\Vert}_{t=1}^k {\rm head}^t({\bmH}^{(l)}_{t})
\end{gather}
\begin{gather}
    {\rm head}^t({\bmH}^{(l)}_{t}) = \bmV^{(l)}_{t}\cdot{\rm softmax}(\frac{\bmK^{(l)\top}_{t}\bmQ^{(l)}_{t}}{\sqrt{d/k}})
\end{gather}
\begin{gather}
    \bmQ^{(l)}_{t} = \bmW^{(l)}_{Q,t}{\bmH}^{(l)}_{t},\ \ \ \  \bmK^{(l)}_{t} = \bmW^{(l)}_{K,t}{\bmH}^{(l)}_{t},\ \ \ \ \bmV^{(l)}_{t} = \bmW^{(l)}_{V,t}{\bmH}^{(l)}_{t},
\end{gather}
where $\bmW_{Q,t}, \bmW_{K,t}, \bmW_{V,t}$ are query, key, and value matrices to be learned by the model, $k$ is the number of attention head and $\Vert$ is the concatenate operation.

\subsection{Problem Formulation}
\noindent\textbf{Definition 2.4. Node Representation Learning on Heterogeneous Text-Rich Networks.} Given a heterogeneous text-rich network $\mathcal{G}=(\mathcal{V}, \mathcal{E}, \mathcal{A}, \mathcal{R}, \mathcal{D})$, the task is to build a model $f_\Theta: \mathcal{V} \rightarrow \mathbb{R}^d$ with parameters $\Theta$ to learn meaningful node representation vectors for both text-rich and \tl nodes, taking heterogeneous network structures and text semantics into consideration. 
The learned node embeddings should be able to generalize to various downstream tasks, such as link prediction, node classification and node clustering.

\section{Methodology}\label{sec::method}
In this section, we present the details of \Ours, the architecture of which is shown in Figure \ref{fig::whole-structure}. 
We first introduce how to conduct text-rich node encoding by jointly considering text information and heterogeneous structure information via a Transformer-based architecture. Then, we illustrate how to perform effective \tl node learning with heterogeneous node type projection and embedding warm-up. Finally, we discuss how to conduct unsupervised model training.

\subsection{Text-Rich Node Encoding}\label{sec::text-rich-encoder}

\subsubsection{Network-aware Node Text Encoding with Virtual Neighbor Tokens}\label{sec::main-model}
Encoding text $doc_{v_i}$ of node $v_i$ in a heterogeneous text-rich network differs from encoding plain text, mainly because node texts are associated with network structure information, which can provide auxiliary signals \cite{hamilton2017inductive}. 
For example, a paper cited by many deep learning papers in an academic network can be highly related to machine learning.
Given that text semantics can be well captured by a multi-layer Transformer architecture \cite{devlin2018bert}, we propose a simple but effective way to inject network signals into the Transformer encoding process. The key idea is to introduce \textit{virtual neighbor tokens}.
Given a node $v_i$ and its associated texts $doc_{v_i}$, let $\bmH^{(l)}_{v_i}\in \mathcal{R}^{d\times n}$ denote the output hidden states of all text tokens in $doc_{v_i}$ after the $l$-th model layer ($l\geq1$). In each layer, we introduce two virtual neighbor tokens to represent $v_i$'s text-rich neighbors $\widehat{N}_{v_i}$ and \tl neighbors $\widecheck{N}_{v_i}$ in the network respectively. Their embeddings are denoted as $\widehat{\bmz}^{(l)}_{v_i}$ and $\widecheck{\bmz}^{(l)}_{v_i} \in \mathcal{R}^d$, which are concatenated to the original text token sequence hidden states as follows (We will discuss how to obtain $\widehat{\bmz}^{(l)}_{v_i}$ and $\widecheck{\bmz}^{(l)}_{v_i}$ in Section \ref{sec::aggregation}.):
\begin{equation}\label{eq::concate}
    \widetilde{\bmH}^{(l)}_{v_i} = \widehat{\bmz}^{(l)}_{v_i} \mathop{\Vert} \bmH^{(l)}_{v_i} \mathop{\Vert} \widecheck{\bmz}^{(l)}_{v_i}.
\end{equation}
After the concatenation, $\widetilde{\bmH}^{(l)}_{v_i}$ contains information from both $v_i$'s accompanied text $doc_{v_i}$ and its neighbors in the network, \textit{i.e.}, $\widehat{N}_{v_i}$ and $\widecheck{N}_{v_i}$. 
(It is worth noting that we insert two virtual neighbor tokens to take the \textit{presence or absence of text heterogeneity} of nodes into consideration.)
To let the text token representations carry network signals, we adopt a multi-head attention mechanism (MHA):
\begin{gather*}
    {\rm MHA}(\bmH^{(l)}_{v_{i}},\widetilde{\bmH}^{(l)}_{v_{i}}) = \mathop{\Vert}_{t=1}^k {\rm head}^t(\bmH^{(l)}_{v_{i},t},\widetilde{\bmH}^{(l)}_{v_{i},t}), \label{mha_asy} \\
    \bmQ^{(l)}_{t} = \bmW^{(l)}_{Q,t}\bmH^{(l)}_{v_{i},t},\ \ \ \  \bmK^{(l)}_{t} = \bmW^{(l)}_{K,t}\widetilde{\bmH}^{(l)}_{v_{i},t},\ \ \ \ \bmV^{(l)}_{t} = \bmW^{(l)}_{V,t}\widetilde{\bmH}^{(l)}_{v_{i},t}.
\end{gather*}
In the equation above, the multi-head attention is asymmetric (\textit{i.e.,} the keys $\bmK$ and values $\bmV$ are augmented with virtual neighbor embeddings but queries $\bmQ$ are not) to avoid network information being overwritten by text signals and utilize refreshed neighbor representations from each Transformer layer. The output of MHA includes updated network-aware representations of text tokens. Then, following the Transformer architecture \cite{vaswani2017attention}, the updated representations will go through a feed-forward network (FFN) to finish our ($l$+1)-th model layer encoding. Formally,
\begin{equation}\label{eq::transformers}
    \begin{gathered}
        \widetilde{\bmH}^{(l)'}_{v_{i}} = {\rm LN}(\bmH^{(l)}_{v_{i}} + {\rm MHA}(\bmH^{(l)}_{v_{i}},\widetilde{\bmH}^{(l)}_{v_{i}})), \\
        \bmH^{(l+1)}_{v_{i}} = {\rm LN}(\widetilde{\bmH}^{(l)'}_{v_{i}} + {\rm MLP}(\widetilde{\bmH}^{(l)'}_{v_{i}})).
    \end{gathered}
\end{equation}
where ${\rm LN}(\cdot)$ denotes the layer normalization function. After $L$ model layers, the final representation of the \cls token will be used as the node representation of $v_i$, \textit{i.e.}, $h_{v_i}=\bmH^{(L+1)}_{v_{i}}\cls$.

\subsubsection{MHA-based Heterogeneous Neighbor Aggregation}\label{sec::aggregation}
The virtual neighbor token embeddings $\widehat{\bmz}^{(l)}_{v_i}, \widecheck{\bmz}^{(l)}_{v_i}$ in Eq.(\ref{eq::concate}) should be the information concentration of $v_i$'s text-rich neighbors and \tl neighbors respectively.
Inspired by the MHA mechanism in Transformer architectures \cite{vaswani2017attention}, we design a multi-head attention module to aggregate the information from neighbors and capture the \textit{type heterogeneity} associated with the \textit{edges}. The neighbor aggregation vector $\bar{\bmz}^{(l)}_{v_i}$ ($\in \{\widehat{\bmz}^{(l)}_{v_i}, \widecheck{\bmz}^{(l)}_{v_i}$\}) of the $l$-th layer for $v_i$ is calculated as follows:
\begin{equation}\label{eq::mha}
    \begin{aligned}
        \bar{\bmz}^{(l)}_{v_i} &= \mathop{\Vert}_{t=1}^k {\rm head}^t(\bmh^{(l)}_{v_i,t}, \{\bmh^{(l)}_{v_j\rightarrow v_i,t}|v_j\in \bar{N}_{v_i}\}), \\
        & = \mathop{\Vert}_{t=1}^k \sum\limits_{v_j \in \bar{N}_{v_i}\cup\{v_i\}} \bar{\alpha}^{(l)}_{{v_i}{v_j},t}\bar{\bmW}^{(l)}_{V,t} \bmh^{(l)}_{v_j\rightarrow v_i,t}.
    \end{aligned}
\end{equation}
In the equation, $\bar{x} \in \{\widehat{x}, \widecheck{x}\}$ ($x$ can be $\bmz_v$, $N_v$, $\alpha$, $W$). $\widehat{x}$ denotes text-rich instances and $\widecheck{x}$ denotes \tl instances. 
 $\bmh^{(l)}_{v_i,t}\in \mathbb{R}^{\frac d k}$ represents the $t$-th chunk of $\bmh^{(l)}_{v_i}$ (For a text-rich node $v_s$, $\bmh^{(l)}_{v_s}=\bmH^{(l)}_{v_{s}}\cls$; For a \tl node $v_p$, we will discuss how to obtain $\bmh^{(l)}_{v_p}$ in Section \ref{sec::textless}.).
 $d$ is the dimension of vectors $\bmh^{(l)}_{v_i}$.
 $\bar{\bmW}_V$ is the value projection matrix.
$\alpha_{v_iv_j}$ is the attention weight of $v_j$ to $v_i$ which is calculated as follows,
\begin{equation}
    \begin{aligned}
        \bar{\alpha}^{(l)}_{{v_i}{v_j},t} &= {\rm softmax}(\bar{e}^{(l)}_{{v_i}{v_j},t}) \\    
        &= {\rm softmax}({\rm Norm}((\bar{\bmW}^{(l)}_{Q,t} \bmh^{(l)}_{x,t} )^\top (\bar{\bmW}^{(l)}_{K,t} \bmh^{(l)}_{{v_j}\rightarrow {v_i},t} ))).
    \end{aligned}
\end{equation}
$\bar{\bmW}_Q$, $\bar{\bmW}_K$ are query and key projection matrices, respectively. $\text{Norm}(\cdot)$ is the scale normalization function, \textit{i.e.}, $\text{Norm}(x)=x/\sqrt{d/k}$. 
$\bmh^{(l)}_{v_j\rightarrow v_i}$ is the propagation vector from $v_j$ to $v_i$, which is calculated as follows depending on the edge type,
\begin{equation}\label{eq::prop}
    \begin{aligned}  
        \bmh^{(l)}_{v_j\rightarrow v_i} = \bmW_r \bmh^{(l)}_{v_j}, \ \text{where}\ \psi(e_{v_jv_i}) = r.
    \end{aligned}
\end{equation} %
 In the equation, $e_{v_jv_i}$ denotes the edge between $v_j$ and $v_i$;
 $\bmW_r$ is the edge type-aware projection matrix, which is designed for capturing the \textit{edge} semantic \textit{heterogeneity}.

\subsection{\Tl Node Encoding}\label{sec::textless}
\subsubsection{Node Type Heterogeneity-based Representation}
In this section, we will discuss how to obtain $\bmh_{v_p}$ for a \tl node $v_p$.
Although they lack semantically rich text, \tl nodes can be quite important as they may contribute significant signals to their neighbors in real-world heterogeneous networks. For example, in an academic network, two papers published in the same venue (textless node) can be on similar topics. Moreover, \tl nodes can also be target nodes for downstream tasks such as author classification.

To align with how we encode text-rich nodes using Transformer-based architectures \cite{vaswani2017attention} (which is introduced in Section \ref{sec::text-rich-encoder}), a straightforward idea of \tl node encoding is to represent each \tl node $v_p$ as a high-dimensional learnable embedding vector $\bmh_{v_p}$ (\textit{e.g.}, $\bmh_{v_p} \in \mathbb{R}^{768}$ to be compatible with BERT-base \cite{devlin2018bert} used in text-rich node encoding). 
Nevertheless, the large population of textless nodes will then introduce a large number of parameters to our framework, which may finally lead to model underfitting.
In addition, due to \textit{node type heterogeneity}, different types of nodes can naturally belong to different latent semantic spaces.
In summary, we design a simple function to calculate $\bmh^{(l)}_{v_p}$ as follows,
\begin{equation}\label{eq::node-proj}
    \bmh^{(l)}_{v_p} = \bmW^{(l)}_{\phi_i} \bmh^{(0)}_{v_p},\ \ \ \text{where}\ \phi(v_p)=\phi_i,\ \ \phi_i\in\mathcal{A}_\text{TL}.
\end{equation}
We set $\bmh^{(0)}_{v_p}$ up as a low-dimensional embedding vector of $v_p$ (\textit{e.g.}, $\bmh^{(0)}_{v_p} \in \mathbb{R}^{64}$) and project it into a more flexible high-dimensional space with a projection matrix $\bmW^{(l)}_{\phi_i}$. 
For different \tl node type $\phi_i\in\mathcal{A}_\text{TL}$, we will have different type-specific projection matrix $\bmW^{(l)}_{\phi_i}$ for the transformation.
By this design, the \textit{node type heterogeneity} is captured.
It is worth noting that the column vectors of $\bmW^{(l)}_{\phi_i}$ can be viewed as ``semantic topic embedding vectors'' \cite{cui2020enhancing} for nodes in type $\phi_i$, and each entry of $\bmh^{(0)}_{v_p}$ represents $v_p$'s weight towards one particular topic. In our experiment, we adopt a shared $\bmW^{(l)}_{\phi_i}$ for each layer, since we find this design can contribute to the best performance.

\subsubsection{\Tl Node Embedding Warm Up}\label{sec::pretrain}
In real-world heterogeneous networks, \tl nodes are of a large population. For example, in academic graphs (\textit{e.g.}, DBLP \cite{tang2008arnetminer}), there are millions of author nodes and thousands of venue nodes, which are not naturally associated with semantically rich texts; in social networks (\textit{e.g.}, Twitter \cite{yang2011patterns}), there are millions of user nodes and hashtag nodes which are \tl.
Learning \tl node embeddings from scratch in Eq. (\ref{eq::node-proj}) seems to be a solution to capture node semantics.
However, the presence of a substantial amount of textless nodes in the framework would add a considerable amount of parameters and may result in the underfitting of the model.
Therefore, the parameter initialization of $\bmh^{(0)}_{v_p}$ and $\bmW^{(l)}_{\phi_i}$ will be crucial towards model optimization \cite{glorot2010understanding}.
To this end, we design a warm-up step to distill information from semantic-rich PLMs into \tl node embeddings, which gives \tl node embeddings good initializations before \Ours finetuning.
The philosophy is to align the initial \tl node embeddings into the same latent space with representations generated by the PLM (which we utilize to initialize the Transformer layers).
The warm-up step is shown below:
\begin{equation}
\label{warm-up}
\small
    \min_{\bmh^{(l)}_{v_p}}\mathcal{L}_{w}=\sum_{\substack{v_p\in\mathcal{V} \\ \phi(v_p)\in \mathcal{A}_\text{TL}}}\sum_{v_u\in \widehat{N}_{v_p}}-\log\frac{\exp(\widebar{\bmh}_{v_u}^{\ \top} \bmh^{(l)}_{v_p})}{\exp(\widebar{\bmh}_{v_u}^{\ \top} \bmh^{(l)}_{v_p})+\sum_{v_u'}\exp(\widebar{\bmh}_{v_u'}^{\ \top} \bmh^{(l)}_{v_p})},
\end{equation}
where $v_u'$ is a text-rich node as a negative sample; $\widebar{\bmh}_{v_u}$ is the output vector (corresponding to the \cls token) from the PLM after encoding the text-rich node $v_u$. Note that parameters in the PLM are fixed here to make this warm-up process efficient. The PLM utilized here should be the same as that loaded for Transformer layers in \Ours (Section \ref{sec::plm}). 
After the warm-up, semantic information from the PLM will be transferred to \tl node embeddings, which will benefit \Ours training a lot.
We will demonstrate the effectiveness of this process in Section \ref{initialization}.

\subsection{Model Training}
\subsubsection{Training Objective}
To train our model, we define the following likelihood objective with parameters $\Theta$:
\begin{equation}\label{likelihood}
    \max_{\Theta} \mathcal{O} = \prod_{\substack{v_i\in \mathcal{V} \\ \phi(v_i)\in\mathcal{A}_\text{TR}}}\prod_{\substack{v_j\in N_{v_i} \\ \phi(v_j)\in\mathcal{A}_\text{TR}}} p(v_j|v_i;\Theta),
\end{equation}
Here, the conditional probability $p(v_j|v_i;\Theta)$ is calculated as follows:
\begin{equation}
    p(v_j|v_i;\Theta) = \frac{\exp(\bmh_{v_j}^\top \bmh_{v_i})}{\sum_{v_u\in \mathcal{V}, \phi(v_u)\in\mathcal{A}_\text{TR}}\exp(\bmh_{v_u}^\top \bmh_{v_i})},
\end{equation}
where $\bmh_{v_i}=\bmh^{(L+1)}_{v_i}$ is the output node embedding generated by \Ours with parameters $\Theta$; $L$ is the number of layers in \Ours.
However, calculating Eq. (\ref{likelihood}) requires enumerating all $(v_j, v_i)$ pairs, which is costly on big graphs. To make the calculation more efficient, we leverage the negative sampling technique \cite{mikolov2013distributed,jin2020multi} to simplify the objective and obtain our loss function below.
\begin{equation}
\label{loss}
\small
    \min_{\Theta} \mathcal{L}=\sum_{\substack{v_i\in \mathcal{V} \\ \phi(v_i)\in\mathcal{A}_\text{TR}}}\sum_{\substack{v_j\in N_{v_i} \\ \phi(v_j)\in\mathcal{A}_\text{TR}}}-\log\frac{\exp(\bmh_{v_j}^\top \bmh_{v_i})}{\exp(\bmh_{v_j}^\top \bmh_{v_i})+\sum_{v_u'}\exp(\bmh_{v_u'}^\top \bmh_{v_i})}.
\end{equation}
In the equation above, $v_u'$ stands for a random negative sample. 
In our implementation, we use ``in-batch negative samples'' \cite{karpukhin2020dense,yang2021graphformers} to reduce the encoding cost.

\subsubsection{Parameter Initialization for Token Embeddings \& Transformer Layers}\label{sec::plm}
It is shown in \cite{erhan2009difficulty} that a good parameter initialization before downstream task fine-tuning is essential for deep learning models. Recently, significant improvements achieved by PLMs \cite{devlin2018bert,liu2019roberta,clark2020electra} in various NLP tasks have also demonstrated this finding. 
In \Ours, a large proportion of parameters in \Ours are token embeddings and parameters in transformer layers. Fortunately, these parameters are well pre-trained in many PLMs \cite{devlin2018bert,liu2019roberta,yang2019xlnet,clark2020electra,brown2020language,radford2019language}. As a result, \Ours can directly load this part of initial parameters from a PLM.

\subsection{Discussions}
\subsubsection{Discussion on the connection between \Ours and GNNs.}
According to Figure \ref{fig::whole-structure}, \Ours adopts a Transformer-based architecture. Meanwhile, it can also be viewed as a graph neural network (GNN) model.
In general, a GNN layer is consisted of neighbor propagation and aggregation to obtain node representations \cite{kipf2017semi, hamilton2017inductive, velivckovic2017graph} as follows,
\begin{small}
\begin{gather}
    \bm{a}^{(l-1)}_{v_iv_j} = {\rm PROP}^{(l)}\left(\bmh^{(l-1)}_{v_i},\bmh^{(l-1)}_{v_j}\right), \big(\forall v_j\in {N}_{v_i}\big); \\ 
    \bmh^{(l)}_{v_i} = {\rm AGG}^{(l)}\left(\bmh^{(l-1)}_{v_i},\{\bm{a}^{(l-1)}_{v_iv_j}|v_j\in {N}_{v_i}\}\right).
\end{gather}
\end{small}
Analogously, in \Ours, Eq. (\ref{eq::prop}) can be treated as the propagation function ${\rm PROP(\cdot)}$, while the aggregation step ${\rm AGG(\cdot)}$ is the combination of Eqs. (\ref{eq::mha}), (\ref{eq::concate}) and (\ref{eq::transformers}). Essentially, the propagation and aggregation function in \Ours are both heterogeneity-aware mechanisms.

\subsubsection{Discussion on Complexity}\label{sec::complexity}
\textbf{Time Complexity:}
Given a center node with $m$ text-rich neighbors (each of which has $p$ tokens) and $n$ \tl neighbors, the time complexity of each \Ours layer's encoding step is $\mathcal{O}(p^2(m+1)+m+n)$, which is on par with the complexity $\mathcal{O}(p^2(m+1))$ of per GNN-cascaded Transformers layer since $m, n\ll p^2m$.
Another straightforward idea of fusing center node text information with its neighbor representations is to directly concatenate token embeddings of the center node, its text-rich neighbors, and its \tl neighbors together and feed them into a PLM. However, in this way, the time complexity of one such layer becomes $\mathcal{O}((p(m+1)+n)^2)$, which is significantly larger than that of our method.
\textbf{Memory Complexity:}
Given a network with $N$ \tl nodes and $T$ parameters in the Transformer layers, the parameter complexity of \Ours is $\mathcal{O}(T+Nd)$, which is the same with heterogeneous GNN-cascaded Transformers \cite{schlichtkrull2018modeling}.


\section{Experiment}

\subsection{Experimental Settings}

\subsubsection{Datasets}
We conduct experiments on three datasets (\textit{i.e.}, DBLP \cite{tang2008arnetminer}, Twitter \cite{zhang2016geoburst}, and Goodreads \cite{wan2018item}) from three different domains (\textit{i.e.}, academic papers, social media posts, and books). 
For DBLP\footnote{\url{https://originalstatic.aminer.cn/misc/dblp.v12.7z}}, we extract papers published from 1990 to 2020 with their author and venue information.
For Twitter\footnote{\url{https://drive.google.com/file/d/0Byrzhr4bOatCRHdmRVZ1YVZqSzA/view?resourcekey=0-3_R5EWrLYjaVuysxPTqe5A}}, we merge the original LA and NY datasets to form a larger dataset. 
For Goodreads\footnote{https://sites.google.com/eng.ucsd.edu/ucsdbookgraph/home}, we remove books without any similar books, and the remaining books with their meta-data fields form the dataset.
The main statistics of the three datasets are summarized in Table \ref{tab:dataset}.

\begin{table}[ht]
\caption{Dataset statistics. *: text-rich node types.}
\vspace{-0.2cm}
\scalebox{0.85}{
\begin{tabular}{c|l|l}
\toprule
Dataset & \multicolumn{1}{c|}{Node} & \multicolumn{1}{c}{Edge} \\
\midrule
\multirow{3}{*}{DBLP} & \# paper*: 3,597,191 & \# paper-paper: 36,787,329 \\
 & \# venue: 28,638 & \# venue-paper: 3,633,613 \\
 & \# author: 2,717,797 & \# author-paper: 10,212,497 \\
\midrule
\multirow{5}{*}{Twitter} & \# tweet*: 279,694 & \multirow{5}{*}{\begin{tabular}[l]{@{}l@{}} \# tweet-POI: 279,694 \\ \# user-tweet: 195,785 \\ \# hashtag-tweet: 194,939 \\ \# mention-tweet: 50,901 \end{tabular}} \\
 & \# POI*: 36,895 & \\
 & \# hashtag: 72,297 & \\
 & \# user: 76,398 & \\
 & \# mention: 24,089 & \\
 \midrule
\multirow{6}{*}{Goodreads} & \# book*:1,097,438 & \# book-book: 11,745,415\\
 & \# shelves: 6,632 & \# shelves-book: 27,599,160 \\
 & \# author: 205,891 & \# author-book: 1,089,145\\
 & \# format: 768  & \# format-book:  588,677\\
 & \# publisher: 62,934 & \# publisher-book: 591,456\\
 & \# language code: 139 & \# language code-book: 485,733\\
\bottomrule            
\end{tabular}
\vspace{-0.2cm}
}
\label{tab:dataset}
\end{table}

\begin{table*}[ht]
\caption{Experiment results on link prediction. *: \Ours significantly outperforms the best baseline with p-value $< 0.05$.}
\setlength{\tabcolsep}{2.5mm}
\scalebox{0.85}{
\begin{tabular}{ll|c|c|c|c|c|c|c|c|c}
\toprule
\multicolumn{2}{c|}{\multirow{2}{*}{Method}} & \multicolumn{3}{c|}{\textbf{DBLP}} & \multicolumn{3}{c|}{\textbf{Twitter}} & \multicolumn{3}{c}{\textbf{Goodreads}} \\
\multicolumn{2}{c|}{} & \textbf{PREC} & \textbf{MRR} & \textbf{NDCG} & \textbf{PREC} & \textbf{MRR} & \textbf{NDCG} & \textbf{PREC} & \textbf{MRR} & \textbf{NDCG} \\ \midrule
\multicolumn{1}{c}{} &MeanSAGE        & 0.7019 &  0.7964 &  0.8437 & 0.6489 & 0.7450 & 0.7991 & 0.6302 & 0.7409 & 0.8001                   \\
\multicolumn{1}{c}{} &BERT        & 0.7569 &  0.8340 &  0.8726 & 0.7179 & 0.7833 & 0.8265 & 0.5571 & 	0.6668 & 0.7395                   \\
\midrule
\multicolumn{1}{c}{\multirow{4}{*}{\rotatebox[origin=c]{90}{Homo GNN} }} & BERT+MeanSAGE   & 0.8131 &  0.8779 &  0.9070 & 0.7201 & 0.7845 & 0.8275 & 0.7301 & 0.8167 & 0.8594                   \\
\multicolumn{1}{c}{} &BERT+MAXSAGE   & 0.8193 &  0.8825	 &  0.9105 & 0.7198 & 0.7845 & 0.8276 & 0.7280 & 0.8164 & 0.8593                   \\
\multicolumn{1}{c}{} &BERT+GAT  &  0.8119  & 0.8771 &  0.9063 &  0.7231 & 0.7873 & 0.8300 & 0.7333 & 0.8170 & 0.8593                 \\
\multicolumn{1}{c}{} &GraphFormers  & 0.8324 &  0.8916 &  0.9175 & 0.7258 & 0.7891 & 0.8312 & 0.7444 & 0.8260 & 0.8665                   \\
\midrule
\multicolumn{1}{c}{\multirow{5}{*}{\rotatebox[origin=c]{90}{Hetero GNN} }} & BERT+RGCN  & 0.7979 &  0.8633 &  0.8945 & 0.7111 & 0.7764 & 0.8209 & 0.7488 & 0.8303 & 0.8699                   \\
\multicolumn{1}{c}{} & BERT+HAN  & 0.8136 &  0.8782 &  0.9072 & 0.7237 & 0.7880 & 0.8306 & 0.7329 & 0.8174 & 0.8597                   \\
\multicolumn{1}{c}{} & BERT+HGT  & 0.8170 &  0.8814 &  0.9098 & 0.7153 & 0.7800 & 0.8237 & 0.7224 & 0.8112 & 0.8552                   \\
\multicolumn{1}{c}{} & BERT+SHGN  & 0.8149 &  0.8785 &  0.9074 & 0.7218 & 0.7866 & 0.8295 & 0.7362 & 0.8195 & 0.8613                   \\
\multicolumn{1}{c}{} & GraphFormers++  & 0.8233 &  0.8856 &  0.9130 & 0.7159 & 0.7799 & 0.8236 & 0.7536 & 0.8328 & 0.8717                   \\
\midrule
\multicolumn{1}{c}{} & \Ours  & \textbf{0.8474*} &  \textbf{0.9019*} &  \textbf{0.9255*} & \textbf{0.7272*} & \textbf{0.7908*} & \textbf{0.8328*} & \textbf{0.7633*} & \textbf{0.8400*} & \textbf{0.8773*}                   \\
\bottomrule
\end{tabular}
}
\label{tab:link-prediction}
\end{table*}

\subsubsection{Baselines}
We compare \Ours with two groups of baselines: \textbf{GNN-cascaded Transformers} and \textbf{Nested Transformers}. The former group can be further classified into \textit{homogeneous GNN-cascaded Transformers}, including BERT+MeanSAGE \cite{hamilton2017inductive}, BERT+Max\\SAGE \cite{hamilton2017inductive} and BERT+GAT \cite{velivckovic2017graph}, and \textit{heterogeneous GNN-cascaded Transformers}, including BERT+RGCN \cite{schlichtkrull2018modeling}, BERT+HAN \cite{wang2019heterogeneous}, BERT+\\HGT \cite{hu2020heterogeneous} and BERT+SHGN \cite{lv2021we}. 
The latter group includes the recent GraphFormers \cite{yang2021graphformers} model. However, GraphFormers can only deal with homogeneous textual networks. To apply it to heterogeneous text-rich networks, we add heterogeneous graph propagation and aggregation in its final layer. This generalized model is named GraphFormers++.
To verify the importance of both text and network information in text-rich networks, we also include vanilla GraphSAGE \cite{hamilton2017inductive} and vanilla BERT \cite{devlin2018bert} in comparison.
Detailed information about the baselines can be found in Appendix \ref{apx:sec:baselines}.

\subsubsection{Reproducibility.}
For all compared models (baselines and \Ours), we adopt the same training objective and the 12-layer BERT-base-uncased \cite{devlin2018bert} as the backbone PLM for a fair comparison. The Adam optimizer \cite{kingma2014adam} with a learning rate 1e-5 and in-batch negative samples with training batch size 30 are used to fine-tune the model. In-batch testing is used for efficiency and the test batch size is 100, 300, and 100 for DBLP, Twitter, and Goodreads, respectively. The maximum length of the PLM is set to be 32, 12, and 64 on the three datasets according to their average document length. For heterogeneous GNN approaches, the embedding size of \tl nodes is 64.
We run experiments on one NVIDIA RTX A6000 GPU.

Following previous studies on network representation learning, we consider three fundamental tasks for quantitative evaluation: link prediction, node classification, and node clustering.

\subsection{Link Prediction}\label{sec::link-prediction}
\noindent\textbf{Settings.}
Link prediction aims to predict missing edges in a network. In order to evaluate the models' ability to encode both text semantics and network structure, we focus on link prediction between two text-rich nodes. Specifically, on DBLP, Twitter, and Goodreads, the prediction is between paper-paper, tweet-POI, and book-book, respectively. The model is trained and tested with in-batch negative sampling and we adopt a 7:1:2 train-dev-test split. Precision@1 (PREC), Mean Reciprocal Rank (MRR), and Normalized Discounted Cumulative Gain (NDCG) are used as evaluation metrics. Given a query node $u$, PREC measures whether the key node $v$ linked with $u$ is ranked the highest in the batch; MRR calculates the average of the reciprocal ranks of $v$; NDCG further takes the order and relative importance of $v$ into account and here we calculate on the full candidate list, the length of which equals to test batch size.

\vspace{3px}
\noindent\textbf{Results.}
Table \ref{tab:link-prediction} shows the performance of all compared methods. 
From Table \ref{tab:link-prediction}, we can observe that: (a) \Ours outperforms all the baseline methods consistently; (b) Transformer+GNN models perform better than both vanilla GNN and vanilla BERT, which demonstrates the importance of encoding both text and network signals in text-rich networks; (c) Network-empowered Transformers including \Ours, GraphFormers, and GraphFormers++ are more powerful than GNN-cascaded Transformers; (d) By considering network heterogeneity, \Ours can have better performance than GraphFormers in heterogeneous text-rich networks.
(e) \Ours yields a larger performance improvement when the network is more dense and heterogeneous (\textit{i.e.}, DBLP, Goodreads vs. Twitter).

\subsection{Node Classification}\label{sec::classification}
\noindent\textbf{Settings.}
In node classification, we train a 2-layer MLP classifier to classify nodes with the generated node embeddings from each model as input. 
The node embeddings are fixed in order to test their representation quality. 
The experiments are conducted on DBLP and Goodreads (because node labels are available in these two datasets) for both \underline{\textbf{text-rich}} and \underline{\textbf{\tl}} nodes. 
For \underline{\textbf{text-rich}} node classification, we focus on paper nodes and book nodes in DBLP and Goodreads, respectively. We select the most frequent 30 classes in DBLP and keep the original 10 classes in Goodreads. Also, we study both \underline{\textit{transductive}} and \underline{\textit{inductive}} node classification to understand the capability of our model comprehensively. For \underline{\textit{transductive}} node classification, the model has seen the classified nodes during representation learning (using the link prediction objective), while for \underline{\textit{inductive}} node classification, the model needs to predict the label of nodes not seen before. 
For \underline{\textbf{\tl}} node classification, we focus on author nodes in both DBLP and Goodreads. The label of each author is obtained by aggregating the labels of his/her publications. We separate the whole dataset into train set, validation set, and test set in 7:1:2 in all cases and each experiment is repeated 5 times in this section with the average performance reported.
Further information can be found in Appendix \ref{apx:classification}.

\vspace{3px}
\noindent\textbf{Results.}
Tables \ref{fig::transductive_text_classification} and \ref{fig::inductive_text_classification} demonstrate the results of different methods in \underline{\textit{transductive}} and \underline{\textit{inductive}} \underline{\textbf{text-rich}} node classification. We observe that: (a) our \Ours outperforms all the baseline methods significantly on both tasks, showing that \Ours can learn more effective node representations for these tasks; (b) Heterogeneous network-based Transformer methods generally achieve better results than homogeneous network-based Transformer methods, which demonstrates the necessity of encoding heterogeneity in heterogeneous text-rich networks; (c) \Ours generalizes quite well on unseen nodes as its performance on inductive node classification is quite close to that on transductive node classification. Moreover, \Ours even achieves higher performance in inductive settings than the baselines do in transductive settings.
Table \ref{fig::textless_classification} reports the result on \underline{\textbf{\tl}} node classification, where we have the following findings: (a) \Ours outperforms all heterogeneous network-based Transformer methods significantly. (b) Compared with text-rich node classification, the improvement of \Ours on \tl node classification over baselines is more significant, indicating that \Ours better captures neighbors' text semantics in \tl node representations.

\begin{table}[t]
\caption{Transductive text-rich node classification.}
\scalebox{0.85}{
\begin{tabular}{c|c|c|c|c}
\toprule
\multirow{2}{*}{Method} & \multicolumn{2}{c|}{\textbf{DBLP}} & \multicolumn{2}{c}{\textbf{Goodreads}} \\
\multicolumn{0}{c|}{} & \textbf{Micro-F1}   & \textbf{Macro-F1}     & \textbf{Micro-F1}   & \textbf{Macro-F1} \\
\midrule
BERT   & 0.6119 &	0.5476 &	0.8364 &	0.7713\\
\midrule
BERT+MaxSAGE   & 0.6179 &	0.5511 &	0.8447 &	0.7866\\
BERT+MeanSAGE     &  0.6198 &	0.5522 &	0.8420 &	0.7826 \\
BERT+GAT    & 0.5943 &	0.5175 &	0.8328 &	0.7713  \\
GraphFormers   & 0.6256 &	0.5616 &	0.8388 &	0.7786\\
\midrule
BERT+HAN     &  0.5965 &	0.5211 &	0.8351 &	0.7747\\
BERT+HGT      &  0.6575 &	0.5951 &	0.8474 &	0.7928  \\
BERT+SHGN     &   0.5982 &	0.5214 &	0.8345 &	0.7737  \\
GraphFormers++     &   0.6474 &	0.5790 &	0.8516 &	0.7993  \\
\midrule
\Ours &  	\textbf{0.6695*} &	\textbf{0.6062*} & \textbf{0.8578*} &	\textbf{0.8076*} \\
\bottomrule
\end{tabular}
}
\label{fig::transductive_text_classification}
\end{table}

\begin{table}[t]
\caption{Inductive text-rich node classification.}
\scalebox{0.85}{
\begin{tabular}{c|c|c|c|c}
\toprule
\multirow{2}{*}{Method} & \multicolumn{2}{c|}{\textbf{DBLP}} & \multicolumn{2}{c}{\textbf{Goodreads}} \\
\multicolumn{0}{c|}{} & \textbf{Micro-F1}   & \textbf{Macro-F1}     & \textbf{Micro-F1}   & \textbf{Macro-F1} \\
\midrule
BERT   & 0.5996 &	0.5318 &	0.8122 &	0.7371\\
\midrule
BERT+MaxSAGE   & 0.6117 &	0.5435 &	0.8368 &	0.7749 \\
BERT+MeanSAGE     &  0.6129 &	0.5431 &	0.8350 &	0.7721 \\
BERT+GAT    & 0.5879 &	0.5150 &	0.8249 &	0.7590  \\
GraphFormers   & 0.6197 &	0.5548 &	0.8330 &	0.7683 \\
\midrule
BERT+HAN     &  0.5948 &	0.5165 &	0.8279 &	0.7626 \\
BERT+HGT      &  0.6467 &	0.5835 &	0.8390 &	0.7798 \\
BERT+SHGN     &   0.5955 &	0.5202 &	0.8280 &	0.7626  \\
GraphFormers++     &   0.6386 &	0.5696 &	0.8427 &	0.7848  \\
\midrule
\Ours &  	\textbf{0.6600*} &	\textbf{0.5976*} &	\textbf{0.8507*} &	\textbf{0.7977*} \\
\bottomrule            
\end{tabular}
}
\label{fig::inductive_text_classification}
\end{table}

\begin{table}[t]
\caption{\Tl node classification.}
\scalebox{0.85}{
\begin{tabular}{c|c|c|c|c}
\toprule
\multirow{2}{*}{Method} & \multicolumn{2}{c|}{\textbf{DBLP}} & \multicolumn{2}{c}{\textbf{Goodreads}} \\
\multicolumn{0}{c|}{} & \textbf{Micro-F1}   & \textbf{Macro-F1}     & \textbf{Micro-F1}   & \textbf{Macro-F1} \\
\midrule
BERT+HAN     &  0.0604 &	0.0270 &	0.4726 &	0.2464 \\
BERT+HGT      &  0.0883 &	0.0539 &	0.4758 &	0.1963 \\
BERT+SHGN     &   0.0619 &	0.0286 &	0.4733 &	0.2457  \\
BERT+RGCN      &  0.2201 &	0.1687 &	0.5768 &	0.3948 \\
GraphFormers++     &   0.1072 &	0.0698 &	0.5007 &	0.2772  \\
\midrule
\Ours & \textbf{0.3817*} &	\textbf{0.3305*} &	\textbf{0.6292*} &	\textbf{0.4835*} \\
\bottomrule            
\end{tabular}
}
\label{fig::textless_classification}
\end{table}

\subsection{Node Clustering}
\noindent\textbf{Settings.}
For node clustering, we utilize KMeans \cite{kanungo2002efficient} to cluster the nodes based on their representations generated by the models. The data and categories used in Section \ref{sec::classification} for text-rich node classification are used here again, but nodes with more than one ground-truth label are filtered. The number of clusters $K$ is set as the number of categories. For DBLP, since the dataset is quite large, we pick the 10 most frequent categories and randomly select 20,000 nodes for efficient evaluation. NMI and ARI \cite{hubert1985comparing} are used as evaluation metrics. Since the performance of KMeans can be affected by the initial centroids, we run each experiment 10 times and report the average performance. In addition to quantitative evaluation, we conduct visualization to depict the distribution of \Ours embeddings, where t-SNE \cite{van2008visualizing} is utilized to project node embeddings into a 2-dimensional space and the nodes are colored based on their ground-truth label.
Further information can be found in Appendix \ref{apx:clustering} and \ref{apx:visualization}.

\begin{figure}[t]
\centering
\subfigure[DBLP]{               
\includegraphics[width=4cm]{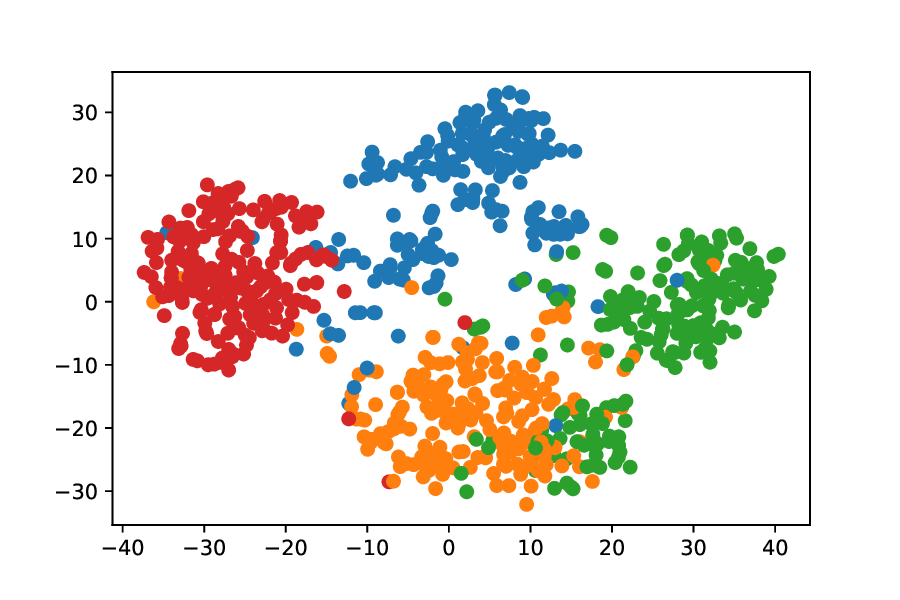}}
\hspace{0in}
\subfigure[Goodreads]{
\includegraphics[width=4cm]{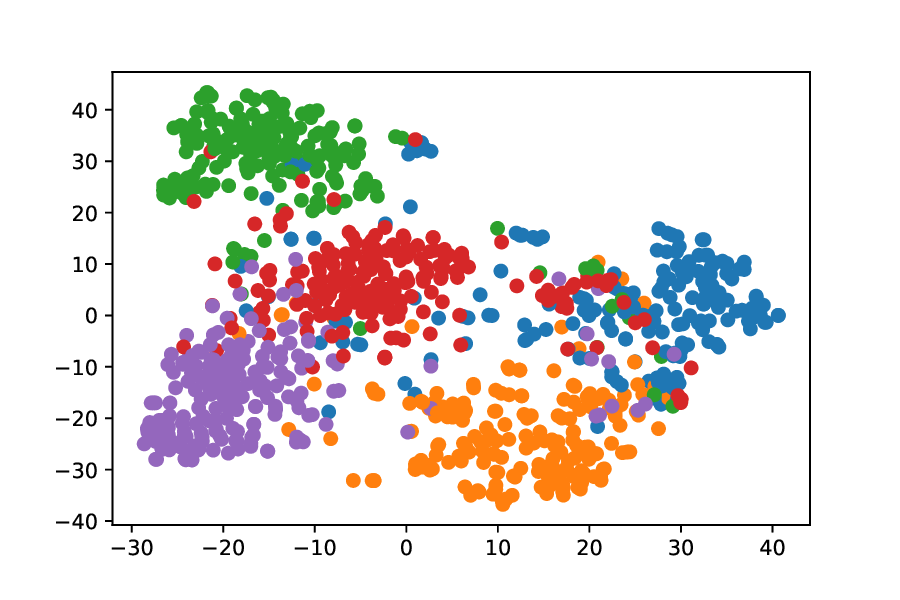}}
\caption{Embedding visualization.}\label{fig::visualization}
\end{figure}

\vspace{3px}
\noindent\textbf{Results.}
The quantitative result can be found in Table \ref{fig::node_clustering}, where \Ours is the best on DBLP and outperforms most baselines on Goodreads. 
The embedding visualization of \Ours is presented in Figure \ref{fig::visualization}. 
In both datasets, the clustering structure is quite evident, indicating that node representations learned by \Ours are category-discriminative, even though the training process is based on link prediction only.

\begin{table}[t]
\caption{Node clustering.}
\scalebox{0.85}{
\begin{tabular}{c|c|c|c|c}
\toprule
\multirow{2}{*}{Method} & \multicolumn{2}{c|}{\textbf{DBLP}} & \multicolumn{2}{c}{\textbf{Goodreads}} \\
\multicolumn{0}{c|}{} & \textbf{NMI}   & \textbf{ARI}     & \textbf{NMI}   & \textbf{ARI} \\
\midrule
BERT   & 0.2570 & 0.3349 & 0.2325 & 0.4013\\
\midrule
BERT+MaxSAGE   & 0.2615 & 0.3490 & 0.2205 & 0.4173\\
BERT+MeanSAGE     &  0.2628 & 0.3488 & \textbf{0.2449} & \textbf{0.4329}\\
BERT+GAT    & 0.2598 & 0.3419 & 0.2408 & 0.4185 \\
GraphFormers   & 0.2633 & 0.3455 & 0.2362 & 0.4139\\
\midrule
BERT+HAN     &  0.2568 & 0.3401 & 0.2391 & 0.4266\\
BERT+HGT      &  0.2469 & 0.3392 & 0.2427 & 0.4296 \\
BERT+SHGN     &   0.2589 & 0.3431 & 0.2373 & 0.4171 \\
GraphFormers++     &   0.2566 & 0.3432 & 0.2372 & 0.4211 \\
\midrule
\Ours &   \textbf{0.2707*} & \textbf{0.3639*} & 0.2429 & 0.4199 \\
\bottomrule            
\end{tabular}
}
\label{fig::node_clustering}
\end{table}

\subsection{Ablation and Parameter Studies}
\subsubsection{Ablation Study on Virtual Neighbor Tokens}
\label{sec:abl-agg}
\begin{figure}[t]
\centering
\subfigure[DBLP]{               
\includegraphics[width=4cm]{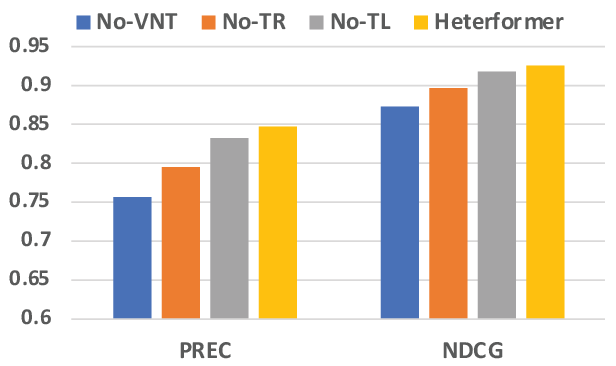}}
\hspace{0in}
\subfigure[Goodreads]{
\includegraphics[width=4cm]{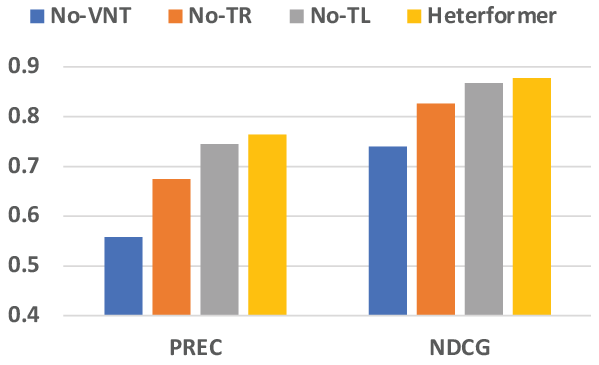}}
\vspace{-0.4cm}
\caption{Ablation study on neighbor aggregation.}\label{fig::ablation}
\vspace{-0.4cm}
\end{figure}

In Section \ref{sec::main-model}, signals from both text-rich and \tl neighbors are incorporated and finally contribute to center node encoding serving as two virtual neighbor tokens. 
To study the effectiveness of information from both text-rich and \tl neighbors, we conduct a model study of several \Ours variants: (a) \textbf{No-VNT} adds no \textbf{V}irtual \textbf{N}eighbor \textbf{T}okens and only encodes textual information for each node; 
(b) \textbf{No-TR} (\textbf{T}ext-\textbf{R}ich) only adds one virtual neighbor token corresponding to \tl neighbors in Eq. (\ref{eq::concate}); 
(c) \textbf{No-TL} (\textbf{T}ext\textbf{L}ess) only adds one virtual neighbor token corresponding to text-rich neighbors in Eq. (\ref{eq::concate}); 
(d) \textbf{\Ours} is our full model. The results of link prediction for these variants are shown in Figure \ref{fig::ablation}. 
We can find that: (a) \Ours outperforms all model variants, which demonstrates that signals from both text-rich and \tl neighbors are essential for center node encoding; (b) No-TL performs better than No-TR, implying that text-rich neighbors are more important than \tl neighbors since they contain rich text semantics.

\subsubsection{Ablation Study on Type Heterogeneous Projection Matrices}
In Eq. (\ref{eq::prop}) and Eq. (\ref{eq::node-proj}), we propose to utilize different projection matrices (so-called heterogeneous projection, \textit{hp} for short) for different types of nodes and edges. In this section, we conduct an ablation study to verify the effectiveness of this design. 
The model with the same projection matrices for different types of nodes and edges is denoted as \textbf{\Ours w/o hp}, while our full model is denoted as \textbf{\Ours}. The results are shown in Table \ref{tab::hp}. From the result, \Ours consistently outperforms \Ours w/o hp on both DBLP and Goodreads, which demonstrates the importance of this design and the necessity of modeling node/edge type heterogeneity.

\begin{table}[t]
\caption{Ablation study on heterogeneous projection.}
\vspace{-0.2cm}
\scalebox{0.85}{
\begin{tabular}{c|c|c|c|c}
\toprule
\multicolumn{1}{c|}{\multirow{2}{*}{Method}} & \multicolumn{2}{c|}{\textbf{DBLP}} & \multicolumn{2}{c}{\textbf{Goodreads}} \\
\multicolumn{1}{c|}{} & \textbf{PREC} & \textbf{MRR} & \textbf{PREC} & \textbf{MRR}  \\
\midrule
\Ours & 0.8474  & 0.9019 & 0.7633 & 0.8400 \\
\midrule
\Ours w/o hp & 0.8415 & 0.8983 & 0.7493 & 0.8325 \\
\bottomrule            
\end{tabular}
}\label{tab::hp}
\end{table}

\subsubsection{Dimension of \Tl Node Embedding}
\begin{figure}[t]
\vspace{-0.3cm}
\centering
\subfigure[DBLP]{               
\includegraphics[width=4cm]{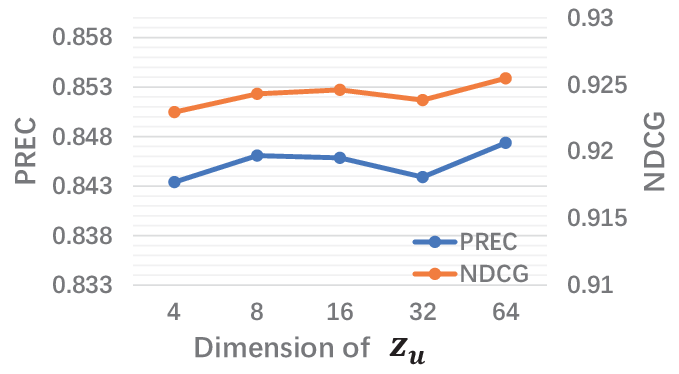}}
\hspace{0in}
\subfigure[Goodreads]{
\includegraphics[width=4cm]{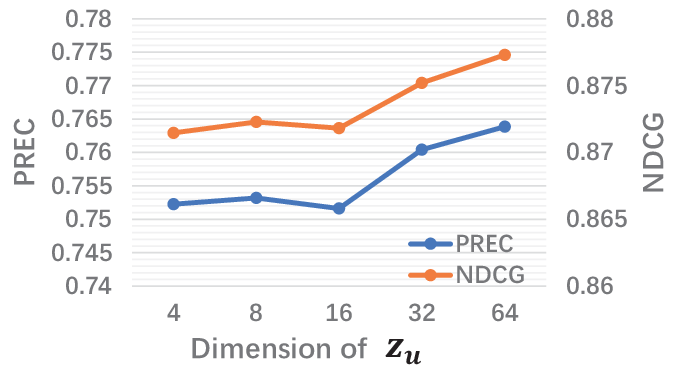}}
\vspace{-0.5cm}
\caption{Effect of \tl node embedding dimension.}\label{fig::hyper} 
\end{figure}

To understand the effect of \tl node embedding dimension, we test the performance of \Ours in link prediction with the embedding dimension varying in 4, 8, 16, 32, and 64. The result is shown in Figure \ref{fig::hyper}. It can be seen that the performance of \Ours generally increases as the embedding dimension becomes larger. This is intuitive since the more parameters $z_u$ has (before overfitting), the more information it can represent.

\subsection{\Tl Node Embedding Warm-Up}\label{initialization}

\subsubsection{Training Curve Study}
\begin{figure}[t]
\centering
\subfigure[DBLP, Validation PREC]{               
\includegraphics[width=4cm]{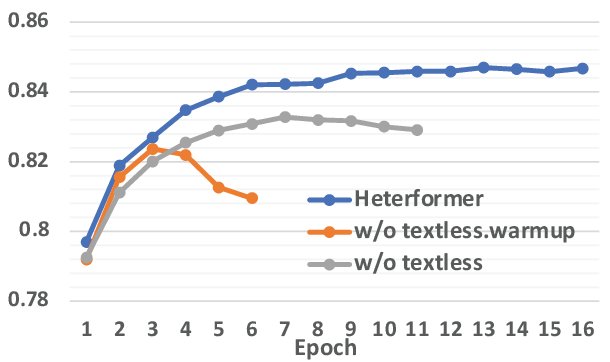}}
\hspace{0in}
\subfigure[Goodreads, Validation PREC]{
\includegraphics[width=4cm]{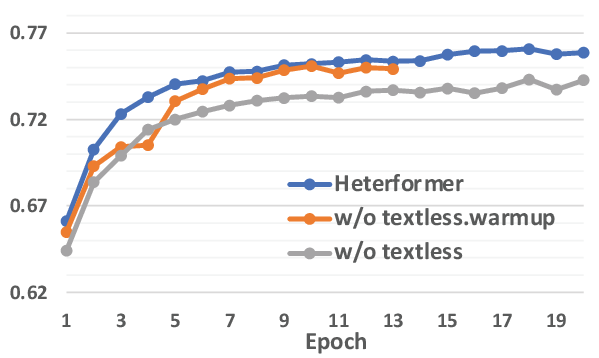}}
\vspace{-0.2cm}
\caption{Performance of \Ours during the training process with and without \tl node warm-up.}\label{fig::initialization}
\vspace{-0.3cm}
\end{figure}

In Section \ref{sec::pretrain}, we propose a method to warm up \tl node embeddings. Now, we empirically demonstrate the effectiveness of such a warm-up process.
The training processes (Section \ref{sec::link-prediction}) of \Ours without \tl node warm-up (\textbf{w/o \tl.warm-up}) and the full \Ours model are shown in Figure \ref{fig::initialization}. We also show \Ours without the utilization of \tl node neighbor information (\textbf{w/o \tl}) as reference. The $x$-axis denotes the number of training epochs, while the $y$-axis represents PREC on the validation set. Training is terminated if PREC on the validation set does not increase for three consecutive epochs.
It is shown that: (a) On both datasets, \Ours with \tl node embedding warm-up can have better performance than that without \tl node warm-up; (b) On DBLP, \Ours without \tl node embedding warm-up cannot even outperform \Ours without the utilization of \tl neighbor information. 
This finding implies the necessity of good initialization for \tl node embeddings. Since modeling \tl nodes can improve the representation capacity (see Section \ref{sec:abl-agg}) in text-rich networks, our warm-up strategy is, therefore, verified to be effective towards model convergence.

\subsubsection{Attention Map Study}
\begin{figure}[t]
\centering
\subfigure[w/o warm-up]{               
\includegraphics[width=4cm]{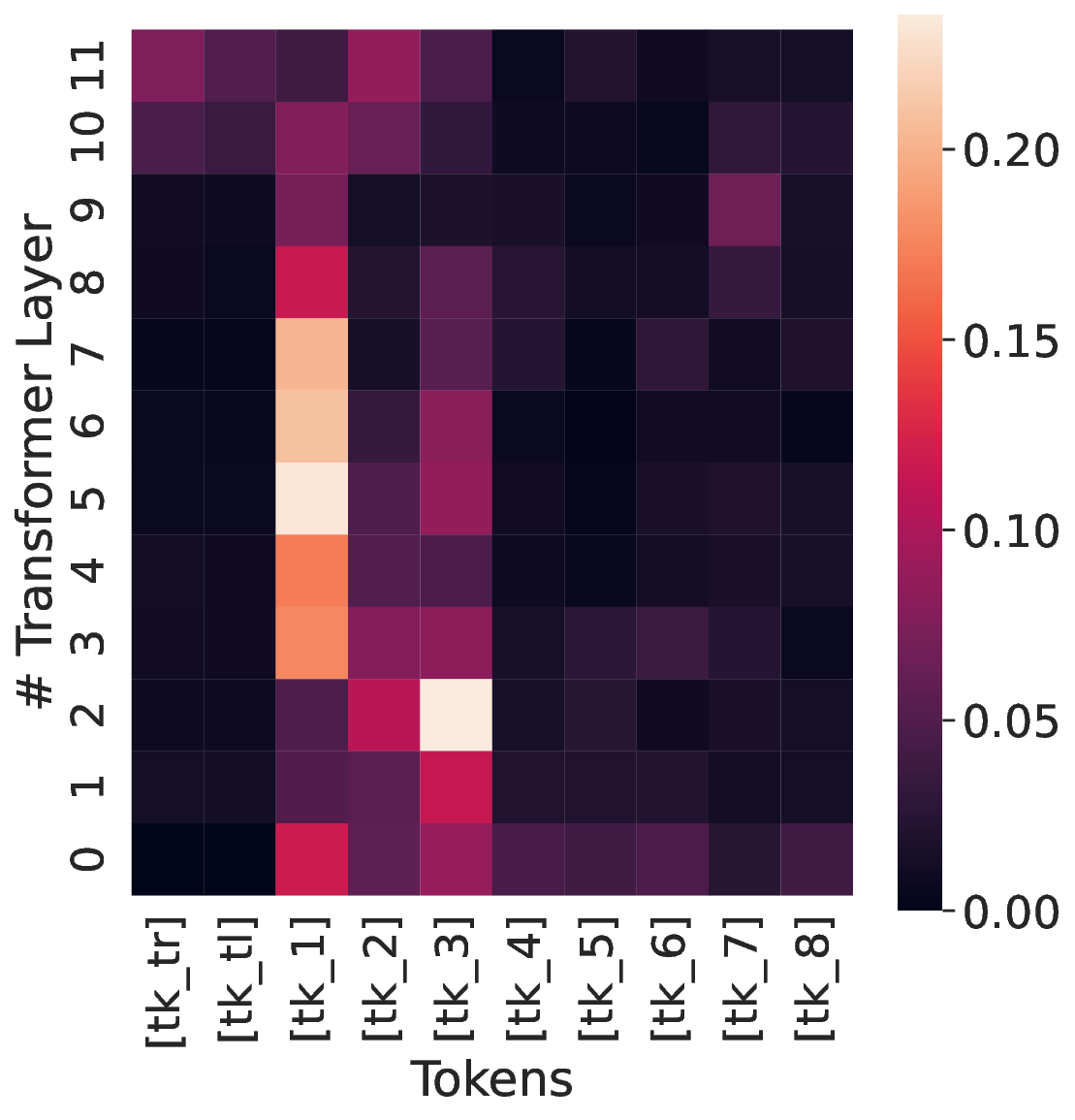}}
\hspace{0in}
\subfigure[w/ warm-up]{
\includegraphics[width=4cm]{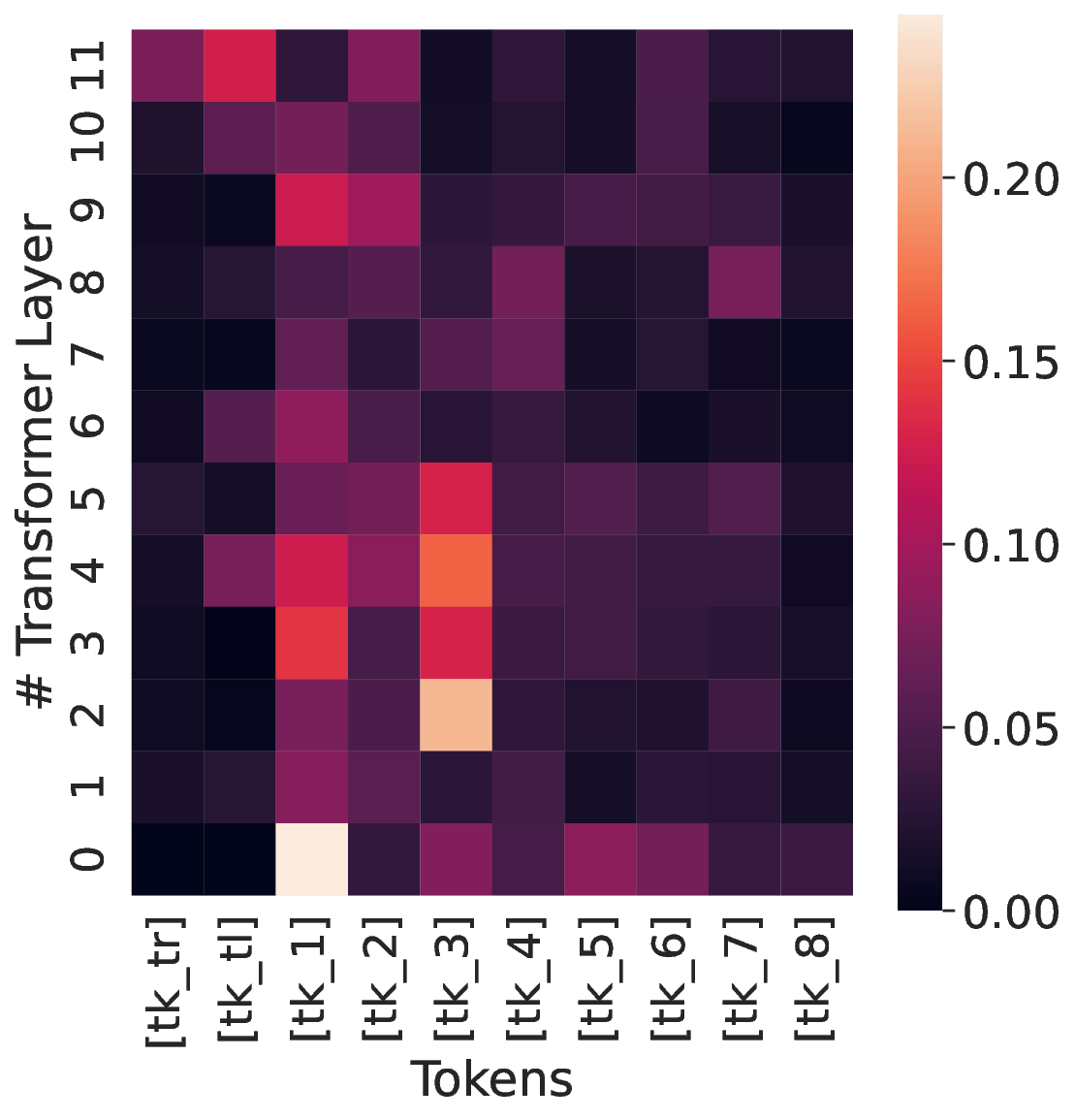}}
\vspace{-0.5cm}
\caption{Self-attention probability map study of \Ours with and without \tl node warm-up for a random sample. The x-axis corresponds to different key/value tokens and the y-axis corresponds to different \Ours layers.}\label{fig::att-map}
\end{figure}

In order to understand how the warm-up step proposed in Section \ref{sec::pretrain} benefits \Ours training, we further conduct a self-attention probability map study for a random sample from DBLP in Figure \ref{fig::att-map}. We random pick up a token from this sample and plot the self-attention probability of how different tokens (x-axis), including virtual neighbor tokens ([tk\_tr] and [tk\_tl] are the text-rich and the \tl neighbor virtual tokens respectively) and the first eight original text tokens ([tk\_x], x$\in\{1..8\}$), will contribute to the encoding of this random token in different layers (y-axis).
From the figure, we can find that the virtual neighbor tokens (first two columns from left) are more deactivated for the model without warm-up, which means that the information from neighbors is not well utilized during encoding. However, the neighbor virtual tokens (first two columns from left) become more activated after warm-up, bringing more useful information from neighbors to enhance center node text encoding.

\subsection{Scalability Study}
We conduct theoretical analysis on time complexity and memory complexity for \Ours in Section \ref{sec::complexity}.
In this section, we perform an empirical time and memory efficiency comparison among BERT+MeanSAGE, GraphFormers, and \Ours. The evaluation is performed on one NVIDIA RTX A6000 GPU. The result is shown in Table \ref{apx:complexity}. For time complexity, we run each model for one mini-batch (each mini-batch contains 30 samples) and report the average running time. For memory complexity, we report the GPU memory needed to train the corresponding models.
From the results, we can find that the time and memory cost of training \Ours is quite close to that of BERT+MeanSAGE and GraphFormers.

\begin{table}[t]
\caption{Scalability Study for BERT+MeanSAGE, Graphformers and \Ours on Goodreads.}
\vspace{-0.2cm}
\scalebox{0.85}{
\begin{tabular}{c|c|c}
\toprule
Model  & Time & Memory \\
\midrule
BERT+MeanSAGE & 440.18ms & 19,637MB \\
\midrule
GraphFormers & 490.27ms  & 20,385MB \\
 \midrule
\Ours & 508.27ms  & 20,803MB \\
\bottomrule            
\end{tabular}
}\label{apx:complexity}
\end{table}

\section{Related Work}

\subsection{Pretrained Language Models}
Pretrained language models (PLMs) aim to learn general language representations from large-scale corpora, which can be generalized to various downstream tasks. Early studies on PLMs mainly focus on context-free text embeddings such as word2vec \cite{mikolov2013distributed} and GloVe \cite{pennington2014glove}. Recently, motivated by the fact that the same word can have different meanings conditioned on different contexts, deep language models such as ELMo \cite{peters2018deep}, BERT \cite{devlin2018bert}, RoBERTa \cite{liu2019roberta}, XLNet \cite{yang2019xlnet}, ELECTRA \cite{clark2020electra}, and GPT \cite{radford2019language,brown2020language} are proposed to capture the contextualized token representations. These models employ the Transformer architecture \cite{vaswani2017attention} to capture long-range and high-order semantic dependency and achieve significant improvement on many downstream NLP tasks \cite{meng2022topic, xun2020correlation, liu2019fine}. However, these models mainly focus on text encoding. In contrast, \Ours leverages both text and heterogeneous structure (network) information when the latter is available.

\subsection{Heterogeneous Graph Neural Networks}
\vspace{-0.1cm}
Graph neural networks (GNNs) such as GCN \cite{kipf2017semi}, GraphSAGE \cite{hamilton2017inductive}, and GAT \cite{velivckovic2017graph} have been widely adopted in representation learning on graphs. Since real-world objects and interactions are often multi-typed, recent studies have considered extending GNNs to heterogeneous graphs \cite{sun2012mining}. The basic idea of heterogeneous graph neural networks (HGNNs) \cite{schlichtkrull2018modeling, wang2019heterogeneous, zhang2019heterogeneous, cen2019representation, yun2019graph, hu2020heterogeneous} is to leverage node types, edge types, and meta-path semantics \cite{sun2011pathsim} in projection and aggregation. For example, HAN \cite{wang2019heterogeneous} proposes a hierarchical attention mechanism to capture both node and meta-path importance; 
HGT \cite{hu2020heterogeneous} proposes an architecture similar to Transformer \cite{vaswani2017attention} to carry out attention on edge types. 
For more HGNN models, one can refer to recent surveys \cite{dong2020heterogeneous,yang2020heterogeneous}. Lv et al. \cite{lv2021we} further perform a benchmark study of 12 HGNNs and propose a simple HGNN model based on GAT. Despite the success of these models, when some types of nodes carry text information, they lack the power of handling textual signals in a contextualized way. In contrast, \Ours jointly models text semantics and heterogeneous structure (network) signal in each Transformer layer.

\subsection{Text-Rich Networks}
\vspace{-0.1cm}
Most previous studies on \textit{homogeneous} text-rich networks adopt a ``cascaded architecture'' \cite{zhu2021textgnn,li2021adsgnn,liu2020fine,jin2021bite,zhou2019gear}.
One drawback of such models is that text and network signals are processed consecutively, so the network information cannot benefit text encoding. 
To overcome this drawback, GraphFormers \cite{yang2021graphformers} introduce nested Transformers so that text and node features can be encoded jointly. 
Edgeformers \cite{jinedgeformers} introduce graph-empowered Transformers for representation learning on textual-edge networks.
However, they assume that the network is homogeneous and all nodes have text information. These assumptions do not usually hold in real-world text-rich networks. Most previous studies on \textit{heterogeneous} text-rich networks focus on specific text-related tasks. For example, HyperMine \cite{shi2019discovering} and NetTaxo \cite{shang2020nettaxo} study how network structures can benefit taxonomy construction from text corpora; LTRN \cite{zhang2021minimally} and MATCH \cite{zhang2021match} leverage document metadata as complementary signals for text classification. In comparison, \Ours focuses on the generic representation learning task.
As far as we know, SHNE \cite{zhang2019shne} is the major previous work also studying representation learning on heterogeneous text-rich networks. However, it still adopts the ``cascaded architecture'' mentioned above and does not explore the power of Transformer encoders (as it was proposed before BERT \cite{devlin2018bert}). In comparison, \Ours proposes a heterogeneous network-empowered Transformer which can jointly capture textual signals and structure signals.

\section{Conclusions}\label{sec::conclusion}
\vspace{-0.1cm}
In this paper, we introduce the problem of node representation learning on heterogeneous text-rich networks and propose \Ours, a heterogeneous network-empowered Transformer architecture to address the problem. 
\Ours can jointly capture the heterogeneous structure (network) information and the rich contextualized textual information hidden inside the networks.
Experimental results on various graph mining tasks, including link prediction, node classification, and node clustering, demonstrate the superiority of \Ours. 
Moreover, the proposed framework can serve as a building block with different task-specific inductive biases. It would be interesting to see its future applications on real-world text-rich networks such as recommendation, abuse detection, tweet-based network analysis, and text-rich social network analysis.

\begin{acks}
This work was supported in part by US DARPA KAIROS Program No. FA8750-19-2-1004 and INCAS Program No. HR001121C0165, National Science Foundation IIS-19-56151, IIS-17-41317, and IIS 17-04532, and the Molecule Maker Lab Institute: An AI Research Institutes program supported by NSF under Award No. 2019897, and the Institute for Geospatial Understanding through an Integrative Discovery Environment (I-GUIDE) by NSF under Award No. 2118329. Any opinions, findings, and conclusions or recommendations expressed herein are those of the authors and do not necessarily represent the views, either expressed or implied, of DARPA or the U.S. Government.
\end{acks}

\bibliographystyle{ACM-Reference-Format}
\bibliography{bib_full_name}

\newpage
\appendix
\clearpage
\section{Supplementary Material}

\subsection{Summary of \Ours's Encoding Procedure}\label{apx::sec::alg}

\RestyleAlgo{ruled}
\SetKwComment{Comment}{/* }{ */}
\begin{algorithm}
\setstretch{0.6}
\SetKwInOut{Input}{Input}\SetKwInOut{Output}{Output}
\caption{Encoding Procedure of \Ours}\label{apx:alg}
\Input{The center node $v_i$, its text-rich neighbors $\widehat{N}_{v_i}$ and \tl neighbors $\widecheck{N}_{v_i}$. Initial token sequence embedding ${\bmH}^{(0)}_{v_j}$ for $v_j\in \widehat{N}_{v_i}\cup \{v_i\}$.}
\Output{The embedding $\bmh_{v_i}$ for the center node $v_i$.}
\Begin{
    \tcp{\textcolor{myblue}{obtain text-rich nodes' first layer encoded token embeddings}}
    \For{$v_j\in\widehat{N}_{v_i}\cup\{v_i\}$}{
        ${\bmH}^{(0)'}_{v_j} \gets {\rm Normalize}(\bmH^{(0)}_{v_j} + {\rm MHA}^{(0)}(\bmH^{(0)}_{v_j}))$ \;
        $\bmH^{(1)}_{v_j} \gets {\rm Normalize}({\bmH}^{(0)'}_{v_j} + {\rm MLP}^{(0)}({\bmH}^{(0)'}_{v_j}))$ \;
        }
        
    \tcp{\textcolor{myblue}{obtain \tl nodes' initial embedding after warm-up}}
    \For{$v_s\in\widecheck{N}_{v_i}$}{
        $\bmh^{(0)}_{v_s} \gets {\rm Warm Up}(v_s)$ \;
        }
    \For{$l=1,...,L$}{
        \tcp{\textcolor{myblue}{text-rich neighbor aggregation}} 
        \For{$v_j\in\widehat{N}_{v_i}\cup\{v_i\}$}{
        ${\bm h}^{(l)}_{v_j} \gets \bmH^{(l)}_{v_j}\cls$ \;
        }
        $\widehat{\bmz}^{(l)}_{v_i} \gets {\rm AGG}(\{{\bm h}^{(l)}_{v_j}|v_j\in \widehat{N}_{v_i}\cup\{v_i\}\})$ \;
        
        \tcp{\textcolor{myblue}{\tl neighbor aggregation}}
        \For{$v_s\in\widecheck{N}_{v_i}$}{
            ${\bm h}^{(l)}_{v_s} \gets \bmW^{(l)}_{\phi_i} \bmh^{(0)}_{v_s}, \ \text{where}\  \phi(v_s)=\phi_i$ \;
            }
        $\widecheck{\bm z}^{(l)}_{v_i} \gets {\rm AGG}(\{{\bm h}^{(l)}_{v_s}|v_s\in \widecheck{N}_{v_i}\cup\{v_i\}\})$ \;
        
        \tcp{\textcolor{myblue}{obtain the center node's token embedding for next layer}}
        $\widetilde{\bmH}^{(l)}_{v_i} \gets \widehat{\bmz}^{(l)}_{v_i} \mathop{\Vert} \bmH^{(l)}_{v_i} \mathop{\Vert} \widecheck{\bmz}^{(l)}_{v_i}$ \;
        $\widetilde{\bmH}^{(l)'}_{v_i} \gets {\rm Normalize}(\bmH^{(l)}_{v_i} + {\rm MHA}^{(l)}(\bmH^{(l)}_{v_{i}},\widetilde{\bmH}^{(l)}_{v_{i}}))$ \;
        $\bmH^{(l+1)}_{v_i} \gets {\rm Normalize}(\widetilde{\bmH}^{(l)'}_{v_i} + {\rm MLP}^{(l)}(\widetilde{\bmH}^{(l)'}_{v_i}))$ \;
        
        \tcp{\textcolor{myblue}{update text-rich neighbors' token embeddings}}
        \For{$v_j\in\widehat{N}_{v_i}$}{
            ${\bmH}^{(l)'}_{v_j} \gets {\rm Normalize}(\bmH^{(l)}_{v_j} + {\rm MHA}^{(l)}(\bmH^{(l)}_{v_j}))$ \;
            $\bmH^{(l+1)}_{v_j} \gets {\rm Normalize}({\bmH}^{(l)'}_{v_j} + {\rm MLP}^{(l)}({\bmH}^{(l)'}_{v_j}))$ \;
            }
        }
    \KwRet{$\bmh_{v_i} \gets \bmH^{(L+1)}_{v_i}\cls$} \;
}
\end{algorithm}

\subsection{Details of Baselines}\label{apx:sec:baselines}
We have 11 baselines including vanilla text/graph encoding models, GNN-cascaded Transformers, and nested Transformers.

\vspace{3px}
\noindent\textbf{Vanilla text/graph models:}
\begin{itemize}[leftmargin=*,partopsep=0pt,topsep=0pt]
    \setlength{\itemsep}{0pt}
    \setlength{\parsep}{0pt}
    \setlength{\parskip}{0pt}
    \item \textbf{MeanSAGE} \cite{hamilton2017inductive}: This is a GNN method utilizing the mean function to aggregate information from neighbors for center node representation learning. 
    The initial node feature vector is bag-of-words weighted by TF-IDF. 
    The number of entries in each attribute vector is the vocabulary size of the corresponding dataset, where we keep the most representative 10000, 2000, and 5000 words for DBLP, Twitter, and Goodreads, respectively, according to the corpora size.
    \item \textbf{BERT} \cite{devlin2018bert}: This is a benchmark PLM pretrained on two tasks: next sentence prediction and mask token prediction. For each text-rich node, we use BERT to encode its text and take the output hidden state of the \cls token as the node representation.
\end{itemize}

\vspace{3px}
\noindent\textbf{Homogeneous GNN-cascaded Transformers:}
\begin{itemize}[leftmargin=*,partopsep=0pt,topsep=0pt]
    \setlength{\itemsep}{0pt}
    \setlength{\parsep}{0pt}
    \setlength{\parskip}{0pt}
    \item \textbf{BERT+MeanSAGE} \cite{hamilton2017inductive}: We stack BERT with MeanSAGE (\textit{i.e.}, using the output text representation of BERT as the input node attribute vector of MeanSAGE). The BERT+MeanSAGE model is trained in an end-to-end way. (Both parameters in BERT and GNN are finetuned.) Other BERT+GNN baselines below have the same cascaded architecture.
    
    \item \textbf{BERT+MaxSAGE} \cite{hamilton2017inductive}: MaxSAGE is a GNN method utilizing the max function for neighbor aggregation to generate center node representation.
    
    \item \textbf{BERT+GAT} \cite{velivckovic2017graph}: GAT is a GNN method with an attention-based neighbor importance calculation, and the importance scores are utilized as weights to aggregate neighbors.
\end{itemize}

\vspace{3px}
\noindent\textbf{Homogeneous Nested Transformers:}
\begin{itemize}[leftmargin=*,partopsep=0pt,topsep=0pt]
    \setlength{\itemsep}{0pt}
    \setlength{\parsep}{0pt}
    \setlength{\parskip}{0pt}
    \item \textbf{GraphFormers} \cite{yang2021graphformers}: This is the state-of-the-art nested Transformer model, which has graph-based propagation and aggregation in each Transformer layer. 
\end{itemize}
Since homogeneous baselines assume all nodes are associated with text information, when applying them to our datasets, we remove all \tl nodes. Therefore, homogeneous baselines cannot be used for \tl node classification (\textit{i.e.}, Table \ref{fig::textless_classification}).

\vspace{3px}
\noindent\textbf{Heterogeneous GNN-cascaded Transformers:}
\begin{itemize}[leftmargin=*,partopsep=0pt,topsep=0pt]
    \setlength{\itemsep}{0pt}
    \setlength{\parsep}{0pt}
    \setlength{\parskip}{0pt}
    \item \textbf{BERT+RGCN} \cite{schlichtkrull2018modeling}: RGCN is a heterogeneous GNN model. It projects neighbor representations into the same latent space according to the edge types. The initial embeddings for \tl nodes are learnable vectors for baselines in this section which is the same to \Ours.
    
    \item \textbf{BERT+HAN} \cite{wang2019heterogeneous}: HAN is a heterogeneous GNN model. 
    It proposes a heterogeneous attention-based method to aggregate neighbor information.
    
    \item \textbf{BERT+HGT} \cite{hu2020heterogeneous}: HGT is a heterogeneous GNN model. Inspired by the Transformer architecture, it utilizes multi-head attention to aggregate neighbor information obtained by heterogeneous message passing.
    
    \item \textbf{BERT+SHGN} \cite{lv2021we}: SHGN is a heterogeneous GNN model. Motivated by the observation that GAT is more powerful than many heterogeneous GNNs \cite{lv2021we}, it adopts GAT as the backbone with enhancements from learnable edge-type embeddings, residual connections, and normalization on the output embeddings.
    
\end{itemize}

\vspace{3px}
\noindent\textbf{Heterogeneous Nested Transformers:}
\begin{itemize}[leftmargin=*,partopsep=0pt,topsep=0pt]
    \setlength{\itemsep}{0pt}
    \setlength{\parsep}{0pt}
    \setlength{\parskip}{0pt}
    \item \textbf{GraphFormers++} \cite{yang2021graphformers}: To apply GraphFormers to heterogeneous text-rich networks, we add heterogeneous graph propagation and aggregation in its final layer. The generalized model is named GraphFormers++.
\end{itemize}

\subsection{Dataset Description}\label{apx:dataset}

\subsubsection{Training}\label{apx:training}
We train our model in an unsupervised way via link prediction. For each paper in DBLP, we select one neighbor paper for it and construct a positive node pair. For each POI in Twitter, we select one neighbor tweet for it to make up a positive node pair. For each book in Goodreads, one neighbor book is selected to build a positive node pair. All these node pairs are used as positive training samples. The model is then trained via in-batch negative sampling. 

\subsubsection{Link Prediction}
The training, validation, and testing sets in this section are the same as those in Section \ref{apx:training}.

\subsubsection{Node classification}\label{apx:classification}

The 30 categories for DBLP papers are: ``Artificial intelligence'', ``Mathematics'', ``Machine learning'', ``Computer vision'', ``Computer network'', ``Mathematical optimization'', ``Pattern recognition'', ``Distributed computing'', ``Data mining'', ``Real-time computing'', ``Algorithm'', ``Control theory'', ``Discrete mathematics'', ``Engineering'', ``Electronic engineering'', ``Theoretical computer science'', ``Combinatorics'', ``Knowledge management'', ``Multimedia'', ``Computer security'', ``World Wide Web'', ``Human-computer interaction'', ``Control engineering'', ``Parallel computing'', ``Information retrieval'', ``Software'', ``Artificial neural network'', ``Communication channel'', ``Simulation'', and ``Natural language processing''. 

The 10 categories for Goodreads books are: ``children'', ``fiction'', ``poetry'', ``young-adult'', ``history, historical fiction, biography'', ``fantasy, paranormal'', ``non-fiction'', ``mystery, thriller, crime'', ``comics, graphic'', and ``romance''.

\subsubsection{Node Clustering}\label{apx:clustering}
For DBLP, since the dataset is quite large, we pick the most frequent 10 categories and randomly select 20,000 nodes for efficient evaluation. The 10 selected categories are: ``Artificial intelligence'', ``Mathematics'', ``Machine learning'', ``Computer vision'', ``Computer network'', ``Mathematical optimization'', ``Pattern recognition'', ``Distributed computing'', ``Data mining'', and ``Real-time computing''.
For Goodreads, we use all 10 categories in the original dataset for clustering.

\subsubsection{Embedding Visualization}\label{apx:visualization}
In this section, we use t-SNE \cite{van2008visualizing} to project node embeddings into low-dimensional spaces. Nodes are colored based on their ground-truth labels. To make the visualization clearer, we select 4 naturally separated categories for DBLP and 5 for Goodreads. The 4 selected categories for DBLP are ``Mathematics'', ``Computer networks'', ``Information retrieval'', and ``Electronic engineering''. The 5 selected categories for Goodreads are ``fiction'', ``romance'', ``mystery, thriller, crime'', ``non-fiction'', and ``children''.

\subsection{Reproducibility Settings}\label{apx:rep}

\subsubsection{Hyper-parameters}
For a fair comparison, the training objective for all compared methods including \Ours and baselines are the same. The hyper-parameter configuration for the node representation learning process can be found in Table \ref{apx:hyper}, where ``neighbor sampling'' means the number of each type of neighbor sampled for the center node during learning.

In Section \ref{sec::classification}, we adopt a multi-layer perceptron (MLP) with 3 layers and hidden dimension 200 to be our classifier. We employ Adam optimizer \cite{kingma2014adam} and early stop 10 to train the classifier. For text-rich node classification, the learning rate is set as 0.001. While for \tl node classification, the learning rate is 0.01.

\begin{table}[t]
\caption{Hyper-parameter configuration.}
\vspace{-0.4cm}
\scalebox{0.8}{
\begin{tabular}{c|c|c|c}
\toprule
Parameter & DBLP & Twitter & Goodreads \\
\midrule
learning rate & \multicolumn{3}{c}{1e-5} \\
\midrule
weight decay & \multicolumn{3}{c}{1e-3} \\
\midrule
adam epsilon & \multicolumn{3}{c}{1e-8} \\
\midrule
early stop & \multicolumn{3}{c}{3} \\
\midrule
\tl embedding & \multicolumn{3}{c}{64} \\
\midrule
chunk $k$ & \multicolumn{3}{c}{12} \\
\midrule
train batch size & \multicolumn{3}{c}{30} \\
\midrule
test batch size & 100 & 300 & 100 \\
\midrule
PLM backbone & \multicolumn{3}{c}{BERT-base-uncased} \\
\midrule
token sequence length & 32 & 12 & 64 \\
\midrule
\multirow{3}{*}{neighbor sampling} & paper:5 & tweet:6 & book:5, shelves:5 \\
& authors:3 & mention:2 & author:2, language code:1  \\
& venue:1 & tag:3,user:1 & publisher:1, format:1 \\

\bottomrule            
\end{tabular}
}\label{apx:hyper}
\end{table}

\subsection{Case Study: Paper Retrieval}
To further demonstrate the capability of \Ours in encoding text semantics, we present a case study of query-based paper retrieval on DBLP. 

\begin{table}[t]
\vspace{-0.3cm}
\caption{Case study of query-based retrieval on DBLP. Top-7 retrieved papers are shown for each method.}
\vspace{-0.4cm}
\scalebox{0.67}{
\begin{tabular}{c|l}
\toprule
\multicolumn{2}{c}{\textbf{Query:} \sethlcolor{yellow}\hl{news recommendation} with \sethlcolor{pink}\hl{personalization}}\\
\midrule
& \multicolumn{1}{c}{\textbf{Retrieved Paper Title}} \\
\midrule
\multirow{7}{*}{\rotatebox[origin=c]{90}{BERT}} & (\ding{55}) \sethlcolor{yellow}\hl{News Recommenders}: Real-Time, Real-Life Experiences \\
\multicolumn{0}{c|}{} & (\ding{55}) \sethlcolor{yellow}\hl{News recommender systems} – Survey and roads ahead \\
\multicolumn{0}{c|}{} & (\ding{55}) A Survey on Challenges and Methods in \sethlcolor{yellow}\hl{News Recommendation} \\
\multicolumn{0}{c|}{} & (\ding{51}) \sethlcolor{pink}\hl{Personalized} \sethlcolor{yellow}\hl{news recommendation}: a review and an experimental investigation \\
\multicolumn{0}{c|}{} & (\ding{51}) Interweaving Trend and User Modeling for \sethlcolor{pink}\hl{Personalized} \sethlcolor{yellow}\hl{News Recommendation} \\
\multicolumn{0}{c|}{} & (\ding{55}) A multi-perspective transparent approach to \sethlcolor{yellow}\hl{news recommendation} \\
\multicolumn{0}{c|}{} & (\ding{55}) Workshop and challenge on \sethlcolor{yellow}\hl{news recommender systems} \\
\midrule
\multirow{7}{*}{\rotatebox[origin=c]{90}{GraphFormers}} & (\ding{51}) \sethlcolor{pink}\hl{Personalized} \sethlcolor{yellow}\hl{news recommendation} based on links of web \\
\multicolumn{0}{c|}{} & (\ding{55}) Interpreting \sethlcolor{yellow}\hl{News Recommendation} Models \\
\multicolumn{0}{c|}{} & (\ding{55}) Do recommendations matter?: \sethlcolor{yellow}\hl{news recommendation} in real life \\
\multicolumn{0}{c|}{} & (\ding{51}) \sethlcolor{pink}\hl{Personalized} \sethlcolor{yellow}\hl{News Recommendation} Based on Collaborative Filtering \\
\multicolumn{0}{c|}{} & (\ding{51}) LOGO: a long-short user interest integration in \sethlcolor{pink}\hl{personalized} \sethlcolor{yellow}\hl{news recommendation} \\
\multicolumn{0}{c|}{} & (\ding{55}) The Intricacies of Time in \sethlcolor{yellow}\hl{News Recommendation} \\
\multicolumn{0}{c|}{} & (\ding{55}) Workshop and challenge on \sethlcolor{yellow}\hl{news recommender systems} \\
\midrule
\multirow{7}{*}{\rotatebox[origin=c]{90}{\Ours}} & (\ding{51}) User attitudes towards \sethlcolor{yellow}\hl{news content} \sethlcolor{pink}\hl{personalization} \\
\multicolumn{0}{c|}{} & (\ding{51}) A system for generating \sethlcolor{pink}\hl{personalized} virtual \sethlcolor{yellow}\hl{news} \\
\multicolumn{0}{c|}{} & (\ding{51}) \sethlcolor{pink}\hl{Personalized} \sethlcolor{yellow}\hl{News Recommendation} Based on Collaborative Filtering \\
\multicolumn{0}{c|}{} & (\ding{55}) Automatic \sethlcolor{yellow}\hl{news recommendations} via aggregated profiling \\
\multicolumn{0}{c|}{} & (\ding{51}) Design and Deployment of a \sethlcolor{pink}\hl{Personalized} \sethlcolor{yellow}\hl{News Service} \\
\multicolumn{0}{c|}{} & (\ding{51}) The design and implementation of \sethlcolor{pink}\hl{personalized} \sethlcolor{yellow}\hl{news recommendation system} \\
\multicolumn{0}{c|}{} & (\ding{51}) \sethlcolor{pink}\hl{Personalizing} \sethlcolor{yellow}\hl{news content}: An experimental study \\
\bottomrule
\end{tabular}
}
\label{tab::retrieve}
\end{table}

\noindent\textbf{Settings.}
The models are asked to retrieve relevant papers for a user-given query based on the inner product of the encoded query embedding and the paper embedding, where the query embedding is obtained by encoding query text only with each model.

\vspace{3px}
\noindent\textbf{Results.}
Table \ref{tab::retrieve} lists the top-7 papers for the query ``\textit{news recommendation with personalization}'' retrieved by BERT, GraphFormers, and \Ours. 
It is shown that our model can have more accurate retrieved results than both baselines. 
In fact, according to network homophily \cite{mcpherson2001birds}, papers on the same topics (\textit{e.g.}, \textit{personalization/news recommendation}) are likely to have connections (\textit{i.e.}, become text-rich neighbors) or share similar meta-data (\textit{i.e.}, share similar \tl neighbors). 
While BERT can consider text information only and GraphFormers enriches text information with text-rich neighbors only, our \Ours is capable of utilizing both text-rich neighbors and \tl neighbors to complement text signals via network-empowered Transformer encoding (Section \ref{sec::text-rich-encoder}), which finally contributes to higher retrieval accuracy.

\end{spacing}

\end{document}